\documentclass{article}

\usepackage{microtype}
\usepackage{graphicx}
\usepackage{subcaption}
\usepackage{booktabs} 

\usepackage{hyperref}
\usepackage{enumitem}

\usepackage[accepted]{icml2026}

\usepackage{amsmath}
\usepackage{amssymb}
\usepackage{mathtools}
\usepackage{amsthm}
\usepackage{multirow}
\usepackage{array}
\usepackage{CJKutf8}

\usepackage[capitalize,noabbrev]{cleveref}

\theoremstyle{plain}

\theoremstyle{definition}

\theoremstyle{remark}

\usepackage[textsize=tiny]{todonotes}

\icmltitlerunning{Self-Prompting Diffusion Transformer for Open-Vocabulary Scene Text Editing via In-Context Learning}

\begin{document}

\twocolumn[
  \icmltitle{Self-Prompting Diffusion Transformer for Open-Vocabulary \\Scene Text Editing via In-Context Learning}



  \icmlsetsymbol{equal}{*}

  \begin{icmlauthorlist}
    \icmlauthor{Hongxi Li}{mt,bit}
    \icmlauthor{Tong Wang}{mt}
    \icmlauthor{Chengjing Wu}{mt}
    \icmlauthor{Tianbao Liu}{mt}
    \icmlauthor{Jiangtao Yao}{mt}
    \icmlauthor{Xiaochao Qu}{mt}
    \icmlauthor{Xinxiao Wu}{bit}
    \icmlauthor{Luoqi Liu}{mt}
    \icmlauthor{Ting Liu}{mt}
  \end{icmlauthorlist}

  \icmlaffiliation{mt}{MT Lab, Meitu Inc., Beijing, China}
  \icmlaffiliation{bit}{School of Computer Science \& Technology, Beijing Institute of Technology, Beijing, China}

  \icmlcorrespondingauthor{Ting Liu}{lt@meitu.com}
  \icmlcorrespondingauthor{Luoqi Liu}{llq5@meitu.com}

  \icmlkeywords{Machine Learning, ICML}

  \vskip 0.3in
]



\printAffiliationsAndNotice{}  

\begin{abstract}
Scene text editing aims to modify text in a target region of an image while preserving surrounding background style and texture.
Existing methods rely solely on image background information while neglecting the visual details of target regions,
which discards stylistic features in the original text and essentially degrades the task to text rendering.
Moreover, the conditions imposed by pre-trained glyph encoder limit the scope of editable text.
To address these issues, this paper proposes a self-prompting scene text editing method that constructs style and glyph prompts directly from the original image, without introducing additional style or glyph encoders.
We employ a two-stage training strategy: the diffusion transformer is first trained on large-scale self-supervised data and then refined using a small set of paired images.
By leveraging the in-context learning capability of the Multi-Modal Diffusion Transformer (MM-DiT), it achieves open-vocabulary and style-consistent text editing.
Experimental results on various languages demonstrate that our method achieves the state-of-the-art performance in both text accuracy and style consistency.
Our project page: \href{https://hongxiii.github.io/mstedit}{hongxiii.github.io/mstedit}.
\end{abstract}

\section{Introduction}
\label{sec:intruduction}
Scene text editing is a specialized yet fundamental image editing task that aims to modify textual content in natural scene images while strictly preserving glyph correctness and visual consistency with the surrounding context. 
By formulating text modification as targeted content generation within localized text regions, scene text editing naturally aligns with image inpainting, while imposing additional constraints on semantic fidelity, typographic structure, and cross-modal consistency.
Driven by recent advances in diffusion-based image inpainting, recent methods~\cite{tuo2023anytext,tuo2024anytext2,zeng2024textctrl, wang2025glyphmastero,wang2025dreamtext} have made notable progress in scene text editing. 
However, these methods inherit the intrinsic limitations of inpainting-based formulations: they struggle to capture fine-grained glyph structures—particularly for logographic scripts and rare languages—and inevitably discard visual information within the original text regions. 
As a result, edited outputs often suffer from character distortion and degraded consistency in style and texture with respect to the source text.

\begin{figure}[t]
    \centering
\includegraphics[width=\columnwidth]{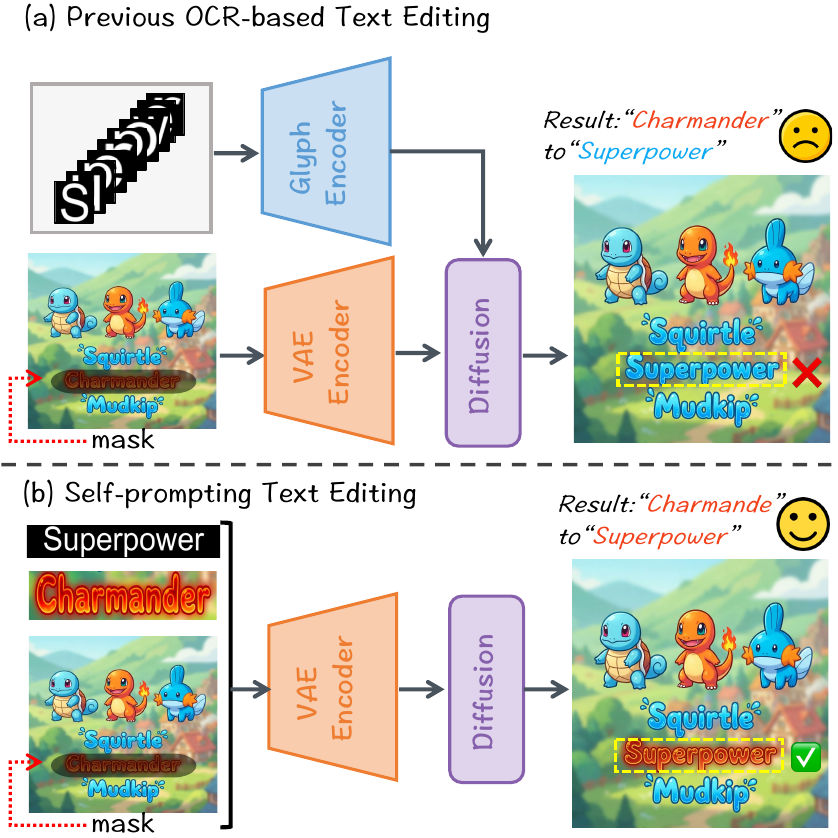}
    \caption{Comparison of previous OCR-based text edit and our proposed self-prompting text edit.}
    \label{fig:introduction}
\end{figure}

The main challenge in scene text editing lies in capturing the glyph features of target text.
An intuitive approach is to utilize a pre-trained optical character recognition (OCR)~\cite{du2020pp} feature extractor as the glyph encoder. However, as OCR systems are inherently designed as classifiers, their fixed vocabularies fundamentally limit the scope of scene text editing, whereas a glyph encoder trained from scratch relies on massive image data.
In addition, existing methods take the design paradigm of image inpainting for text editing, which discards the original information of the target region. As a consequence, the generated text often fails to preserve the pre-editing style and instead borrows stylistic cues from surrounding regions.
As illustrated in Figure~\ref{fig:introduction}(a), although previous methods have modified the red word "Charmander" to the target word "Superpower", the target text retains the same blue font as the surrounding text, thereby losing its original stylistic texture.

To address these issues, this paper proposes a self-prompting open-vocabulary text editing method. We capture glyph features at the stroke level rather than the character level, without the need for additional encoders. 
We leverage the in-context learning capability of the Multi-Modal Diffusion Transformer (MMDiT)~\cite{flux2024} to  to enhance the model’s generalization performance. This design enables the model to adapt to complex real-world scenarios and support diverse languages with only a small-scale paired dataset, while simultaneously ensuring strict style consistency between the original and edited text.
As demonstrated in Figure~\ref{fig:introduction}(b), our method successfully modifies the red word "Charmander" to the red word "Superpower". Notably, the edited text retains the original color, font, and texture of the target region, even though the surrounding text in the background remains blue.

Specifically, for input prompt construction, we generate a high-fidelity glyph map by rendering the target text, which serves as the glyph prompt to guide text content generation. Concurrently, we extract the original pixel information from the target text region in the input image to form a style prompt, which encodes the unique color, texture, and font style of the original text. These two prompts are then concatenated with the full input image to form a multi-modal input tensor for the MMDiT backbone.
Second, we freeze the model’s encoder and decoder components and exclusively train the backbone using large-scale self-supervised image-text datasets. This step equips the model with fundamental text inpainting capabilities without overfitting to specific text styles.
Finally, we utilize a mask-free image editing tool to collect and filter high-quality paired image datasets, where each pair consists of an original image and its corresponding style-consistent edited version. These curated datasets are then used for the cooldown training, during which the model learns to align the generated text with the original style of the target region.

The main contributions of our proposed self-prompting scene text edit can be summarized as follow:
\begin{itemize}
    \item We achieve open-vocabulary text editing by in-context learning to capture stroke-level features from rendered glyph images, rather than character-level features.
    \item We enhance pre- and post-editing style consistency of the target text via cooldown training on a limited amount of paired data, without introducing additional style encoders.
    \item Extensive experiments on the AnyWord-3M and MST-Edit datasets demonstrate that our method outperforms existing methods in text accuracy and style fidelity across 13 evaluated languages.
\end{itemize}

\section{Related Work}
\label{sec:related work}

\subsection{Image Inpainting}
\label{subsec:image inpainting}
Image Inpainting aims to fill missing image regions with visually coherent and semantically plausible content while remaining consistent with surrounding context.
Early methods rely on patch-based propagation and GAN-based encoder–decoder architectures with structural priors such as partial or gated convolutions and contextual attention~\cite{liu2018image,yu2018generative,yu2019free}.
More recent advances are dominated by diffusion-based approaches, which can be categorized into sampling-based methods that modify the denoising process in a training-free manner~\cite{avrahami2022blended,lugmayr2022repaint}, and fine-tuning-based methods that explicitly encode masks and masked images to improve content and shape awareness~\cite{rombach2022high,manukyan2023hd,zhuang2024task}.
To balance performance and generality, recent plug-and-play designs decouple masked image conditioning from generation, enabling effective inpainting without retraining entire diffusion backbones~\cite{zhang2023adding,ju2024brushnet}.

Image inpainting serves as the foundation of scene text editing, where accurate background restoration is crucial for consistent text synthesis.

\subsection{Scene Text Editing}
\label{subsec:scene text editing}
Scene text editing extends image inpainting by jointly requiring background restoration and geometry- and style-consistent text synthesis.
Early methods~\cite{roy2020stefann,yang2020swaptext,qu2023exploring} adopt multi-stage GAN-based pipelines with explicit geometric or stroke priors, offering controllable edits but suffering from poor generalization and error accumulation in complex scenes.
Recent methods~\cite{tuo2023anytext,tuo2024anytext2,wang2025dreamtext}employ pre-trained OCR models as glyph encoders to capture character-level structures. While this design improves glyph awareness, it inherently restricts generalization to characters outside the fixed OCR vocabulary. In contrast, FluxText~\cite{lan2025flux} and TextFlux~\cite{xie2025textflux} remove glyph encoders and instead render glyphs directly within masked regions to extract visual features; however, the spatial extent of the target region can hinder the accurate capture fine-grained glyph details. 
Despite their differences, these methods largely adhere to an image inpainting formulation that discards the original text region, leading to the loss of pre-existing style information and degraded stylistic consistency during editing.
TextCtrl~\cite{zeng2024textctrl} employs a text style encoder to capture color, font, texture, and background features from the original text. However, the representation mismatch between the glyph-structure encoder and the VAE image latent space restricts the method’s scalability beyond isolated text regions.

Differing from existing methods, our method directly constructs style and glyph prompts from the original image without additional encoders, thereby enabling open-vocabulary and style-consistent text editing.

\section{Method}
\label{sec:method}

\subsection{Preliminary}
\label{subsec:preliminary}
Our method is built upon \emph{FLUX-Fill}~\cite{flux2024}, an inpainting-oriented variant of MMDiT, which formulates image editing as a conditional rectified flow process in the latent space and employs a transformer-based architecture to jointly reason over visual content, spatial structure, and textual semantics under multimodal conditions.

\textbf{Masked Image Construction.}
Given an input image $I \in \mathbb{R}^{H \times W \times 3}$ and a binary mask $M \in \{0,1\}^{H \times W}$ indicating the target text region, FLUX-Fill first constructs a masked image
\begin{equation}
I_m = I \odot (1 - M),
\end{equation}
where $\odot$ denotes element-wise multiplication.
The masked image $I_m$ preserves the visual context outside the editable region while removing the original content within the target area.
Both the original image $I$ and the masked image $I_m$ are encoded into latent representations using a frozen VAE encoder, producing visual tokens that serve as the input to the diffusion transformer.
The binary mask $M$ is also embedded and provided to the model as an explicit spatial prior, enabling region-aware generation during denoising.

\textbf{Text Prompt Encoding.}
FLUX-Fill employs two complementary text encoders to extract semantic guidance from textual prompts.
Specifically, a T5 encoder is used to process the full natural-language prompt, capturing high-level semantic intent and contextual information.
The resulting embeddings are injected into the diffusion transformer through cross-attention layers, guiding global content generation and scene-level consistency.
In parallel, a CLIP text encoder is used to encode concise textual descriptors that emphasize visual alignment, such as object names or short phrases.
CLIP embeddings primarily serve to enhance vision--language alignment and stabilize the correspondence between generated content and visual context.

\textbf{Dual-Stream and Single-Stream Transformer.}
The core of FLUX-Fill is a hybrid transformer design that alternates between dual-stream and single-stream blocks.
In the dual-stream transformer blocks, visual tokens and text tokens are processed in separate streams with independent self-attention operations.
Cross-attention is then applied to enable information exchange between modalities, allowing the model to align textual semantics with spatially grounded visual representations while maintaining modality-specific feature structures.

After cross-modal interaction, the architecture transitions to single-stream transformer blocks, where visual and text tokens are concatenated into a unified token sequence and processed jointly.

\textbf{Rectified Flow Objective.}
Let $z_0$ denote the clean latent representation obtained from the VAE encoder and $z_1 \sim \mathcal{N}(0, I)$ be Gaussian noise.
A noisy latent $z_t$ is constructed via rectified flow interpolation at timestep $t$, and the model is trained to predict the velocity field connecting $z_t$ and $z_0$:
\begin{equation}
\mathcal{L}_{\text{RF}} =
\mathbb{E}_{t, z_0, z_1}
\left[
\left\|
\hat{v}_\theta(z_t, t, c) - (z_1 - z_0)
\right\|_2^2
\right],
\label{equa:RF loss}
\end{equation}
where $c$ denotes the multimodal conditioning inputs, including visual tokens, text embeddings, and spatial mask information.

\begin{figure*}[htbp]
    \centering
    \includegraphics[width=\textwidth]{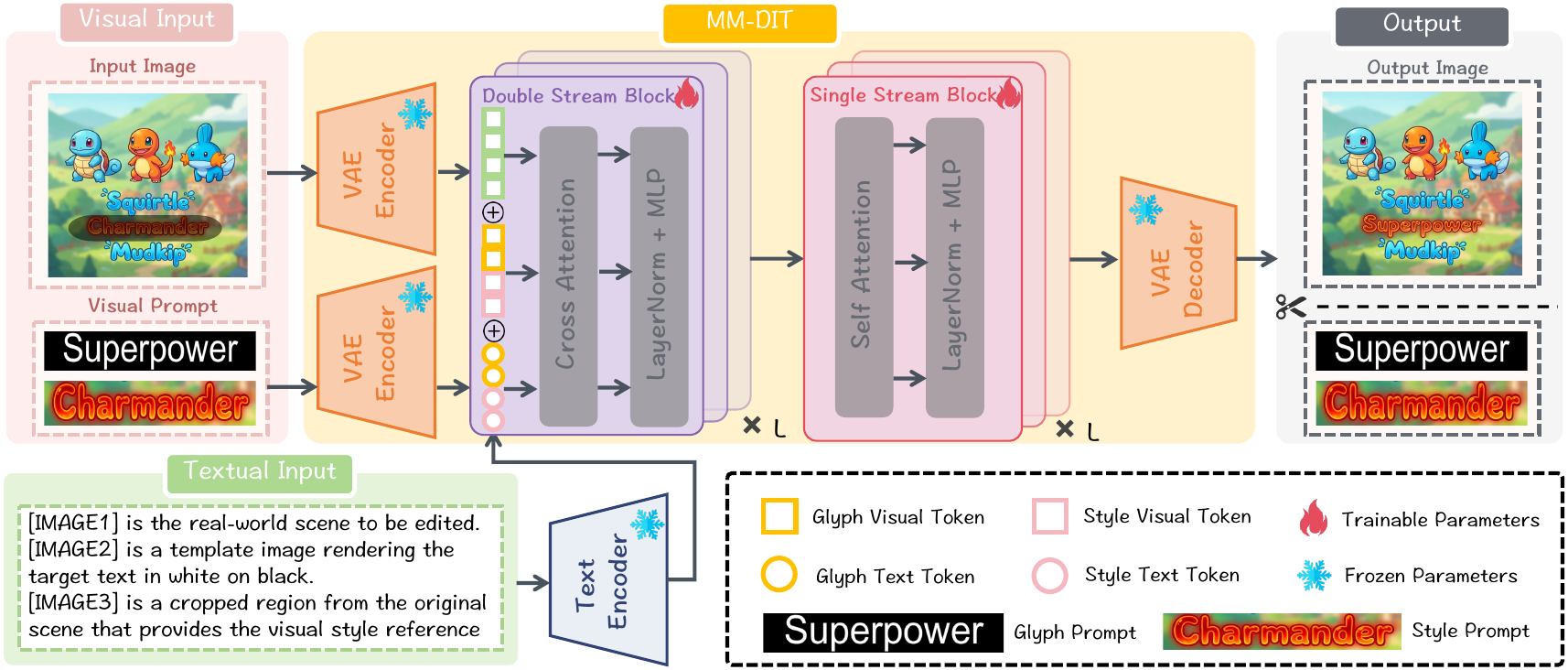}
    \caption{Overview of our proposed method.}
    \label{fig:framework}
\end{figure*}

\subsection{Overall Architecture}
\label{subsec:overall architecture}

Our method explicitly disentangles text style and text content through dedicated visual and textual prompts, while fully leveraging the in-context learning capability of the MMDiT backbone.
An overview of the proposed architecture is shown in Figure~\ref{fig:framework}.

\textbf{Style Prompt Construction.}
To preserve the visual appearance of the original text, we construct a \emph{style prompt} from both visual and textual perspectives.

Given an input image $I \in \mathbb{R}^{H \times W \times 3}$ and a binary mask
$M \in \{0,1\}^{H \times W}$ indicating the target text region,
we compute the maximal enclosing bounding rectangle of the masked area and crop the corresponding region from $I$ to obtain the visual style prompt $I_s$.
This cropped patch encodes region-specific appearance information, such as color, texture, font characteristics, and local illumination, and serves as a visual reference for style preservation.

For style text prompt, we encode the input text description using the CLIP text encoder. Together, the visual and textual style prompts enable the model to infer and preserve the intrinsic style of the target text region without introducing additional style-specific encoders.

\textbf{Glyph Prompt Construction.}
To guide the generation of target text content, we construct a \emph{glyph prompt} that explicitly represents the desired textual structure.

Specifically, the target text is rendered into a single-line glyph image using the Pillow library, producing a white-on-black glyph map $I_g$.
This high-contrast representation preserves fine-grained stroke-level geometry and provides explicit structural guidance for complex glyphs across different languages and scripts, reducing the burden on the diffusion model to learn character structures from scratch.

For glyph text prompt, we encode the target text string using the T5 text encoder.
The resulting embeddings capture high-level semantic and syntactic information of the target text and are injected into the MMDiT backbone through cross-attention, guiding content-aware generation.

\textbf{Denoising Process.}
The visual glyph prompt $I_g$, visual style prompt $I_s$, and masked image $I_m$ are concatenated along the channel dimension to form the composite visual input \[
I_{\text{input}} = \mathrm{Concat}(I_g, I_s, I_m)
\]

The composite input is encoded by a frozen VAE encoder $\mathcal{E}(\cdot)$ to obtain the latent representation $z_0$, which is processed by the MMDiT backbone together with the textual glyph and style embeddings.

After the denoising process, the predicted latent representation is decoded by the VAE decoder $\mathcal{D}(\cdot)$ to produce an edited image.
Since the masked image is used only for conditioning, the final output is obtained by cropping the decoded result to the spatial region corresponding to the target text area.

\subsection{Cooldown Training}
\label{subsec:cooldown training}

Existing OCR-based datasets discard original text appearance by masking target regions, providing only self-supervised rendering signals and limiting style preservation.
To enable style-consistent text editing under limited paired data, we adopt a two-stage cooldown training strategy.

We first construct a paired image dataset tailored for style-aware learning. Specifically, we leverage an instruction-based image editing model (\emph{Nano Banana Pro}~\footnote{\url{https://nanobanana.im/nano-banana-pro}}) to generate edited images conditioned on explicit editing instructions. Each data pair consists of an original image and a corresponding edited image, in which only the target text region is modified while all other regions are strictly preserved.
To ensure data reliability, we manually filter the generated pairs and retain only samples that satisfy the following criteria:
\begin{itemize}
    \item Non-target regions remain unchanged.
    \item The generated text content is semantically accurate.
    \item The edited text preserves consistent style, color, and texture with respect to the original text.
\end{itemize}

\textbf{Self-supervised Pretraining.} We first pretrain the model on the AnyWord-3M dataset, which provides large-scale self-supervised data for multilingual scene text rendering.
During this stage, the optimization objective follows Eq.~\ref{equa:RF loss}, where the conditioning includes multimodal visual and textual inputs. 

\begin{figure}[!h]
    \centering
    \includegraphics[width=\columnwidth]{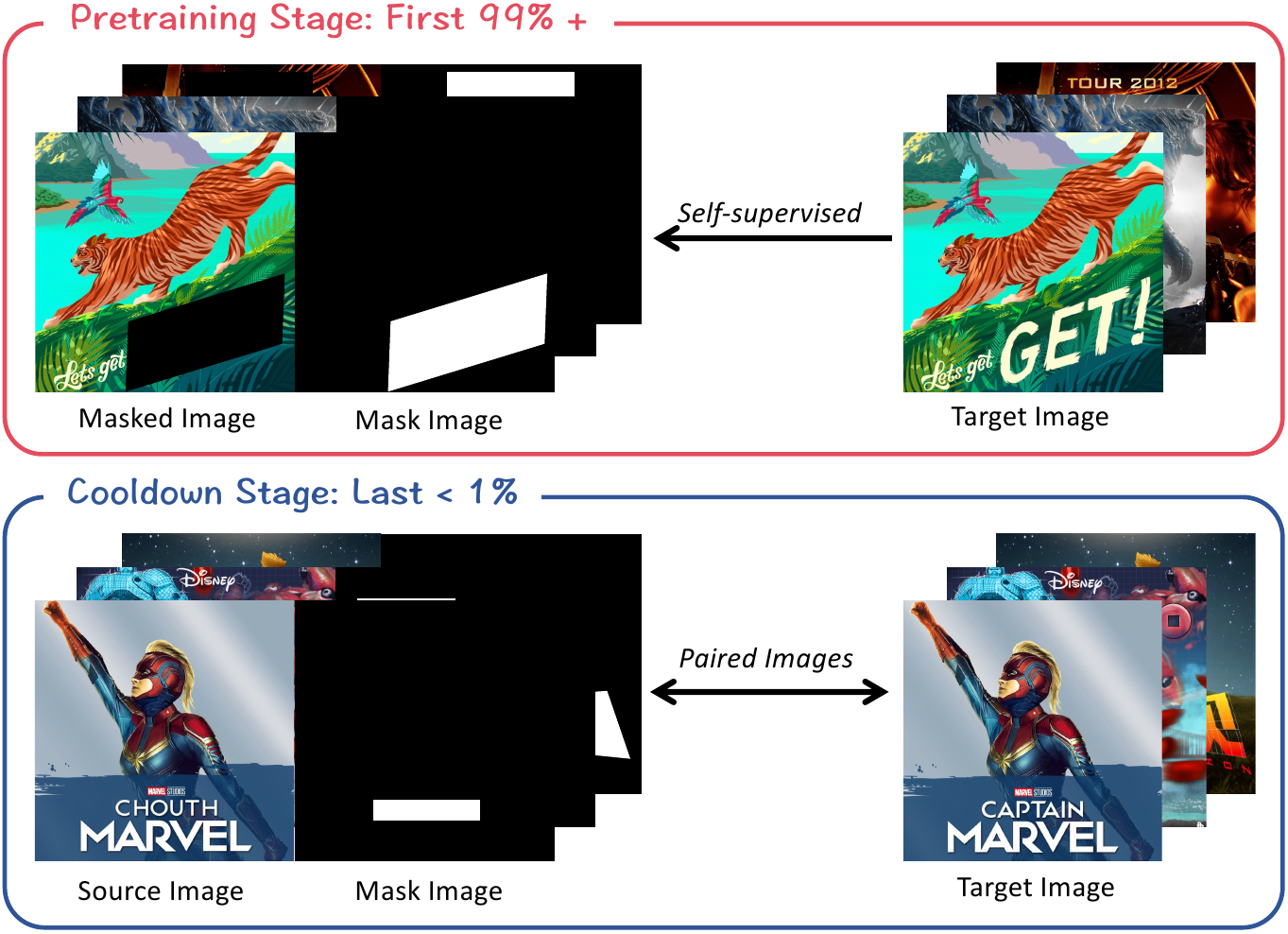}
    \caption{A comparison between data used by standard pre-training and cooldown training.}
    \label{fig:cooldown}
\end{figure}

\textbf{Cooldown Training.}
Inspired by MECO~\cite{gao2025metadata}, we treat the original text region in the input image as meta-information during training.
We continue training from the pretrained checkpoint using a curated paired dataset consisting of 4,000 manually filtered image pairs.
Each pair contains an original image and a corresponding edited image with style-consistent text replacement. Detailed construction process are provided in Appendix~\ref{sup:dataset}. The comparison between data used by pre-training and cooldown stage is shown in Figure~\ref{fig:cooldown}

To mitigate the degenerate optimization where original text region features unduly dominate the target region (leading the model to replicate both original style and glyphs), we design a target-region-oriented training objective for the cooldown stage. This objective restricts the learning signal to localized text transformations, prompting the model to decouple style preservation from content regeneration in optimization.

Let $z_0^{\mathrm{src}}$ and $z_0^{\mathrm{tgt}}$ denote the latent representations of the source image and its corresponding edited image, respectively.
We construct an interpolated latent variable
\begin{equation}
z_t = (1 - \sigma_t)\, z_0^{\mathrm{src}} + \sigma_t\, z_0^{\mathrm{tgt}},
\end{equation}
where $\sigma_t \in [0,1]$ is a time-dependent interpolation coefficient.
The model is trained to predict the velocity field that connects the source and target latents by minimizing
\begin{equation}
\mathcal{L}_{\mathrm{CD}} =
\mathbb{E}_{t}
\left[
\left\|
\hat{v}_\theta(z_t, t, c) -
\left(z_0^{\mathrm{tgt}} - z_0^{\mathrm{src}}\right)
\right\|_2^2
\right],
\end{equation}
where the conditioning $c$ includes the original text region as meta-information through the style prompt.

\section{Experiment}
\label{sec:experiment}

\subsection{Experimental Setup}
\label{subsec:experimental setup}
\textbf{Datasets.} We adopt AnyWord-3M~\cite{tuo2023anytext} as the large-scale benchmark dataset. Its training set contains 1.6M Chinese images, 1.39M English images, and 10K images spanning Japanese, Korean, Arabic, Bengali, and Hindi.
The test split includes 1,000 Chinese and 1,000 English images, and is denoted as AnyText-benchmark.

\begin{table}[!t]
\centering
\caption{Composition of the MST-Edit dataset.}
\label{tab:mst edit}
\begin{tabular}{l l r}
\hline
\textbf{Source Dataset} & \textbf{Language} & \textbf{\# Images} \\
\hline
\multirow{8}{*}{ICDAR-19}
& Arabic   & 1,000 \\
& French   & 1,000 \\
& German   & 1,000 \\
& Korean   & 1,000 \\
& Japanese & 1,000 \\
& Italian  & 1,000 \\
& Bengali  & 1,000 \\
& Hindi    & 1,000 \\
\hline
RusTitW
& Russian  & 3,795 \\
\hline
ThaiOCRBench
& Thai     & 2,808 \\
\hline
Swahili-STR
& Swahili  & 985 \\
\hline
\end{tabular}
\end{table}

We further construct a Multi-lingual Scene Text Editing Dataset (MST-Edit) by aggregating multiple publicly available multilingual datasets, including ICDAR-19~\cite{nayef2019icdar2019}, RusTitW~\cite{markov2023rustitw}, ThaiOCRBench~\cite{nonesung2025thaiocrbench}, and Swahili-STR~\cite{douamba2024first}. Detailed statistics of the dataset composition are summarized in Table~\ref{tab:mst edit}.
We randomly split 20\% of MST-Edit for testing, with the remainder used for training.

\textbf{Evaluation metrics.} For textual accuracy, Sentence Accuracy (Seq. ACC) quantifies the correctness of generated text at the sentence level, and Normalized Edit Distance (NED) measures character-level similarity between the generated and target text. For visual fidelity, Fréchet Inception Distance (FID)~\cite{Seitzer2020FID} quantifies the distribution alignment of generated and real images in the Inception-v3 feature space, and Learned Perceptual Image Patch Similarity (LPIPS)~\cite{zhang2018unreasonable} measures the L2 distances between perceptual VGG-based deep features.

\begin{table*}[!t]
\centering
\caption{Quantitative results on AnyText-benchmark. \textbf{Bold} indicates the best result and \underline{underline} indicates the second best.}
\resizebox{\linewidth}{!}{
\begin{tabular}{lcccccccc}
\toprule
\multirow{2}{*}{\textbf{Method}} & \multicolumn{4}{c}{\textbf{English}} & \multicolumn{4}{c}{\textbf{Chinese}} \\
\cmidrule(lr){2-5} \cmidrule(lr){6-9}
 & Sen. Acc($\uparrow$) & NED($\uparrow$) & FID($\downarrow$) & LPIPS($\downarrow$) & Sen. Acc($\uparrow$) & NED($\uparrow$) & FID($\downarrow$) & LPIPS($\downarrow$) \\
\midrule
TextDiffuser~\cite{chen2023textdiffuser} & 0.5176 & 0.7618 & 29.76 & 0.1564 & 0.0559 & 0.1218 & 34.19 & 0.1252 \\
AnyText~\cite{tuo2023anytext} & 0.6843 & 0.8588 & 21.59 & 0.1106 & 0.6476 & 0.8210 & 20.01 & 0.0943 \\
TextCtrl~\cite{zeng2024textctrl} & 0.5853 & 0.8146 & 35.73 & 0.1978 & 0.3580 & 0.6084 & 49.79 & 0.2298 \\
AnyText2~\cite{tuo2024anytext2} & 0.7915 & 0.9100 & 29.76 & 0.1734 & 0.7022 & 0.8420 & 26.52 & 0.1444 \\
GlyphMastero~\cite{wang2025glyphmastero} & 0.8170 & - & - & - & 0.7301 & - & - & - \\
FIUX-Text~\cite{lan2025flux} & 0.8175 & 0.9193 & \underline{12.35} & \underline{0.0674} & 0.7213 & 0.8555 & \underline{12.41} & \underline{0.0487} \\
TextFlux~\cite{xie2025textflux} & \underline{0.8231} & \underline{0.9235} & 13.42 & 0.0721 & \underline{0.7289} & \underline{0.8612} & 13.67 & 0.0524 \\
Our method 
& \textbf{0.8857} & \textbf{0.9568} & \textbf{7.62} & \textbf{0.0365} 
& \textbf{0.8249} & \textbf{0.9147} & \textbf{7.95} & \textbf{0.0268} \\
\bottomrule
\end{tabular}
}
\label{tab:anytext-bench result}
\end{table*}

\textbf{Implementation Details.}
All experiments are conducted on a cluster with 8 NVIDIA A100 GPUs.
Our model is initialized from FLUX.1-Fill-Dev\footnote{\url{https://huggingface.co/black-forest-labs/FLUX.1-Fill-dev}}.
During training, we freeze the VAE, CLIP text encoder, and T5 text encoder, and update only the transformer parameters.
We follow the default FLUX configuration, using a fixed guidance scale of 30 and 30 sampling steps for both training and inference.

Training is conducted in two stages. We first train the model on AnyWord-3M for one epoch, followed by a 10-epoch cooldown phase on the paired image dataset. Multilingual data are jointly mixed without language-specific scheduling. We use the AdamW optimizer with a constant learning rate of $2\times10^{-5}$, bf16 mixed precision, and 8-bit optimizer states. The per-GPU batch size is 1 with gradient accumulation over 8 steps, resulting in an effective batch size of 64, and all experiments are conducted with a fixed random seed of 42.

\subsection{Quantitative Result}
\label{subsec:quantitative result}

We perform quantitative experiments under both bilingual (Chinese/English) and multilingual (non-Chinese/English) settings.
For Chinese and English, we evaluate our method against state-of-the-art methods~\cite{chen2023textdiffuser,tuo2023anytext,tuo2024anytext2,zeng2024textctrl,wang2025glyphmastero,xie2025textflux,lan2025flux} on the AnyText-benchmark.

For non-Chinese/English languages, we re-implement two OCR-free baselines~\cite{xie2025textflux,lan2025flux} and compare them with our method on the MST-Edit dataset, which covers the remaining 11 languages.

Table~\ref{tab:anytext-bench result} reports the quantitative results on the AnyText benchmark. Based on these results, we make the following observations:

\noindent(1) \textbf{Overall performance.}
Our method achieves the best performance across all metrics on both the English and Chinese subsets, consistently outperforming existing methods.

\noindent(2) \textbf{Text accuracy.}
Compared with the second-best method, TextFlux, our approach improves sentence-level accuracy from 0.8231 to 0.8857 on English and from 0.7289 to 0.8249 on Chinese. The NED score is also increased from 0.9235 to 0.9568 and from 0.8612 to 0.9147, respectively, indicating improved preservation of fine-grained glyph structures.

\noindent(3) \textbf{Image quality.}
More significant gains are observed on image quality metrics. Our method reduces FID to 7.62/7.95 and LPIPS to 0.0365/0.0268 on English/Chinese, surpassing the strongest baselines and demonstrating superior visual fidelity and stylistic consistency.

\begin{figure}[!t]
    \centering
    \includegraphics[width=\columnwidth]{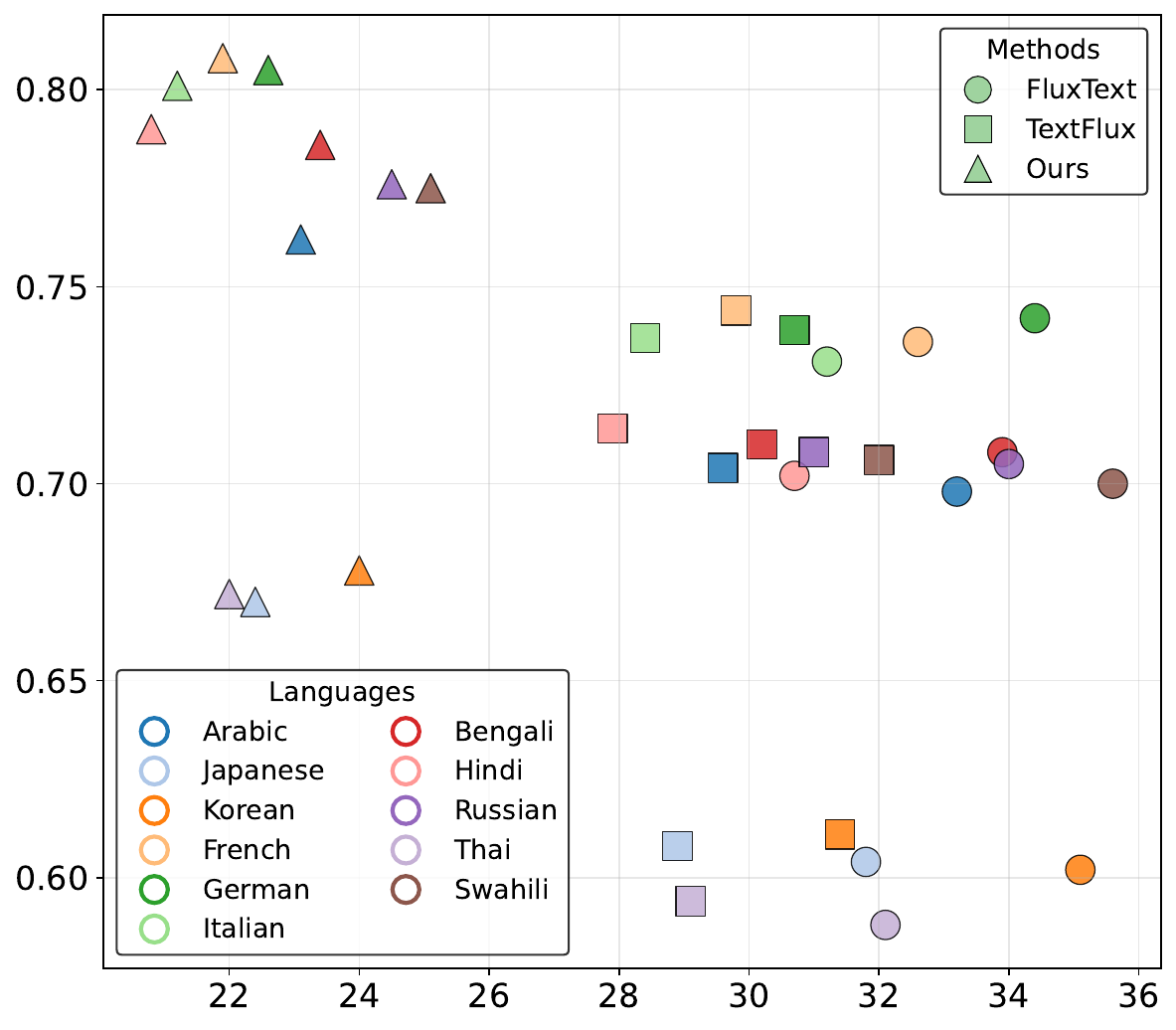}
    \caption{Quantitative comparison of OCR-free methods on MST-Edit, with Seq. ACC on the Y-axis and FID on the X-axis.}
    \label{fig:mst-edit result}
    \vspace{-4mm}
\end{figure}
The experimental results on MST-Edit are illustrated in Figure~\ref{fig:mst-edit result}. Overall, our method consistently outperforms the two OCR-free baselines across all evaluated languages, demonstrating a clear method-level advantage. From a cross-lingual perspective, languages belonging to the Latin script family (e.g., French, German, and Italian) generally achieve better performance, which can be attributed to their shared alphabetic structures and similar stroke patterns. In contrast, languages with more complex glyph compositions and diverse stroke layouts, such as Thai, tend to exhibit relatively lower performance. These results suggest that character structural complexity plays an important role in multilingual scene text editing, while our method remains robust across diverse writing systems. Complete results on MST-Edit are provided in the Appendix~\ref{sup:mstedit}.

\subsection{Ablation Study}
\label{subsec:ablation study}

We conduct ablation studies to analyze the effect of style prompts and to investigate potential cross-lingual negative interference in multilingual scene text editing.

\begin{table}[!h]
\centering
\caption{Ablation study on the effect of the cooldown stage.}
\label{tab:ablation-style}
\begin{tabular}{llcc}
\toprule
~ & \textbf{Metric} & \textbf{w/o Cooldown} & \textbf{w/ Cooldown} \\
\midrule
\multirow{4}{*}{\rotatebox{90}{English}}
& Sen. Acc ($\uparrow$) & 0.8738 & \textbf{0.8857} \\
& NED ($\uparrow$)      & 0.9470 & \textbf{0.9568} \\
& FID ($\downarrow$)    & 11.56 & \textbf{7.62} \\
& LPIPS ($\downarrow$)  & 0.0608 & \textbf{0.0365} \\
\midrule
\multirow{4}{*}{\rotatebox{90}{Chinese}}
& Sen. Acc ($\uparrow$) & 0.8125 & \textbf{0.8249} \\
& NED ($\uparrow$)      & 0.8906 & \textbf{0.9147} \\
& FID ($\downarrow$)    & 12.01 & \textbf{7.95} \\
& LPIPS ($\downarrow$)  & 0.0425 & \textbf{0.0268} \\
\bottomrule
\end{tabular}
\label{tab:ablation style}
\end{table}

\begin{figure}[!h]
    \centering
    \includegraphics[width=\columnwidth]{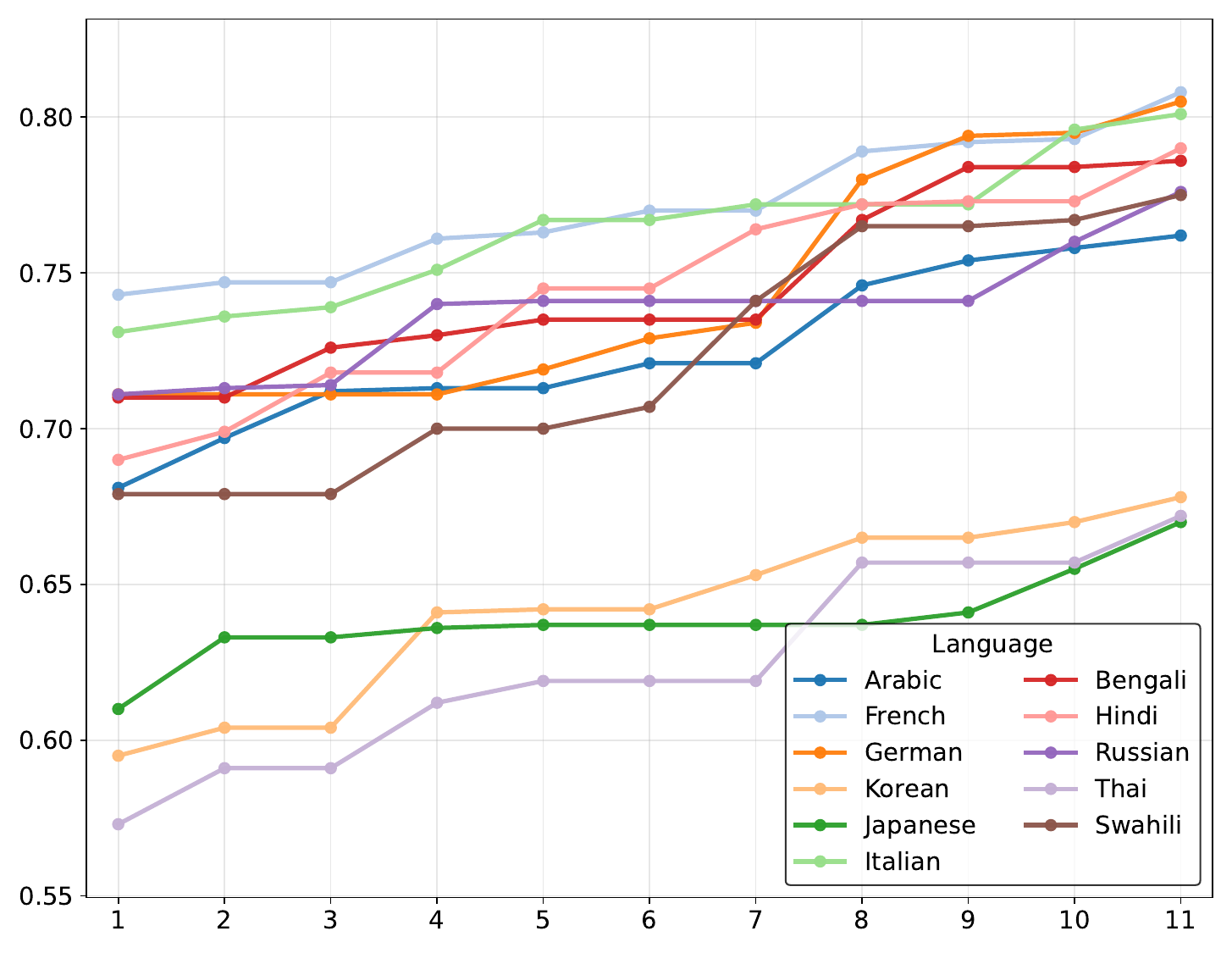}
    \caption{Effect of the number of introduced languages on Seq. ACC of initially introduced languages.}
    \label{fig:ablation language}
\end{figure}

To analyze the impact of style prompts, we conduct an ablation study by removing visual style prompts from the image input and textual style prompts from the text input, while disabling the cooldown training described in Section~\ref{subsec:cooldown training}.

Results on the AnyText benchmark are reported in Table~\ref{tab:ablation style}. 
With the cooldown stage enabled, FID is reduced from 11.56 to 7.62 on English and from 12.01 to 7.95 on Chinese, while LPIPS decreases from 0.0608 to 0.0365 and from 0.0425 to 0.0268, respectively. In contrast, improvements in text accuracy are relatively modest. These results indicate that style prompts, reinforced by the cooldown stage, primarily improve image quality, while offering complementary gains in text accuracy.

We investigate the impact of progressively incorporating additional language data on earlier-introduced languages using a cyclic training protocol, in which languages are sequentially added in the order \textit{{Arabic, English, French, Chinese, German, Korean, Japanese, Italian, Bengali, Hindi, Russian, Thai, Swahili}}.

Empirically, Seq. ACC remains stable and exhibits a consistent upward trend as additional languages are introduced, as illustrated in Figure~\ref{fig:ablation language}. This behavior is consistent with our design principle of learning stroke-level visual primitives rather than language- and character-specific representations: shared low-level stroke structures across writing systems are reinforced by multilingual exposure, leading to improved generalization. Complete numerical results are provided in the Appendix~\ref{sup:mstedit}.

\begin{figure}[!t]
    \centering
    \includegraphics[width=\columnwidth]{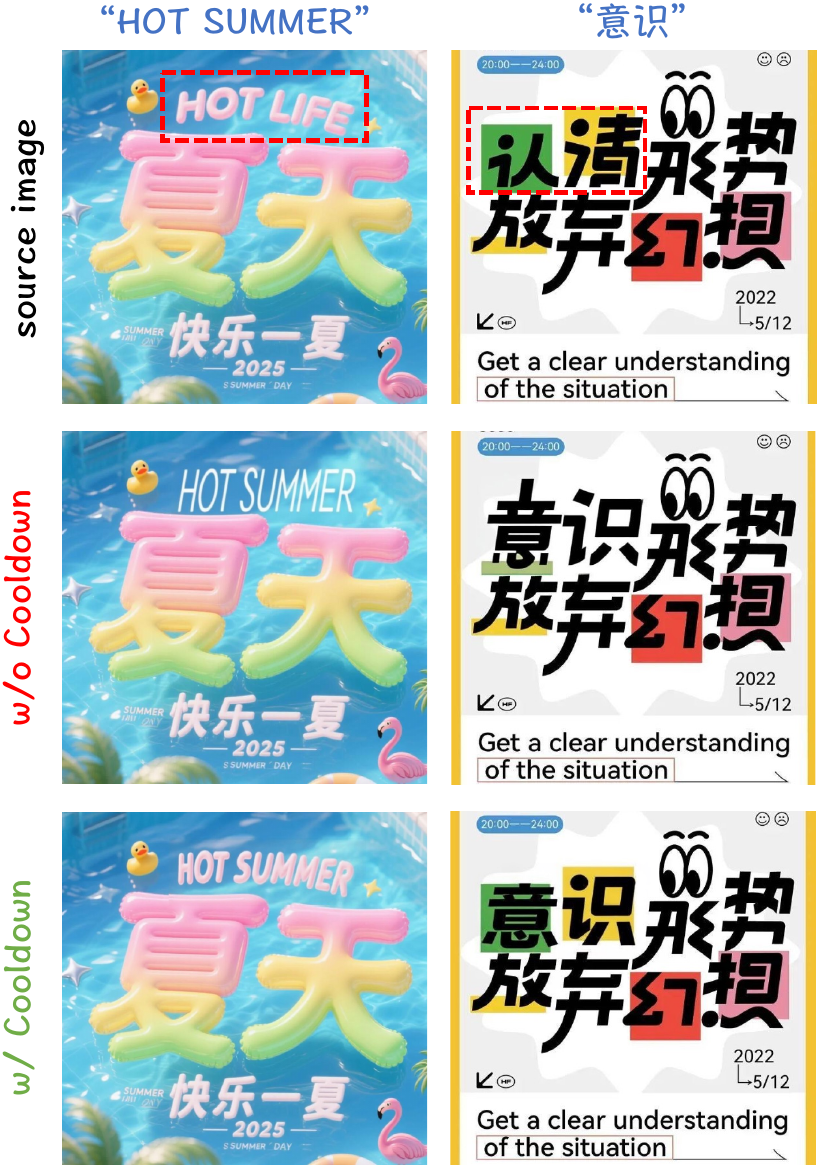}
    \caption{Comparison of scene text edit results with and without style prompts.}
    \label{fig:ablation example}
    \vspace{-4mm}
\end{figure}
Figure~\ref{fig:ablation example} presents qualitative image editing results with and without style prompts. It can be observed that both settings accurately generate the target text without erroneous strokes. However, in the left example, when the style prompt is removed, the target text “HOT SUMMER” is rendered using a font style that does not appear in the original image, leading to noticeable stylistic inconsistency. In the right example, although the result without style prompts renders the target text ``\begin{CJK}{UTF8}{gbsn}意识\end{CJK}'' using a font consistent with the surrounding text, it fails to preserve the original text background.
In contrast, the setting with style prompts produces target text that more faithfully restores both the original font style and background, resulting in higher visual fidelity.

\begin{figure*}[!h]
    \centering
    \includegraphics[width=\textwidth]{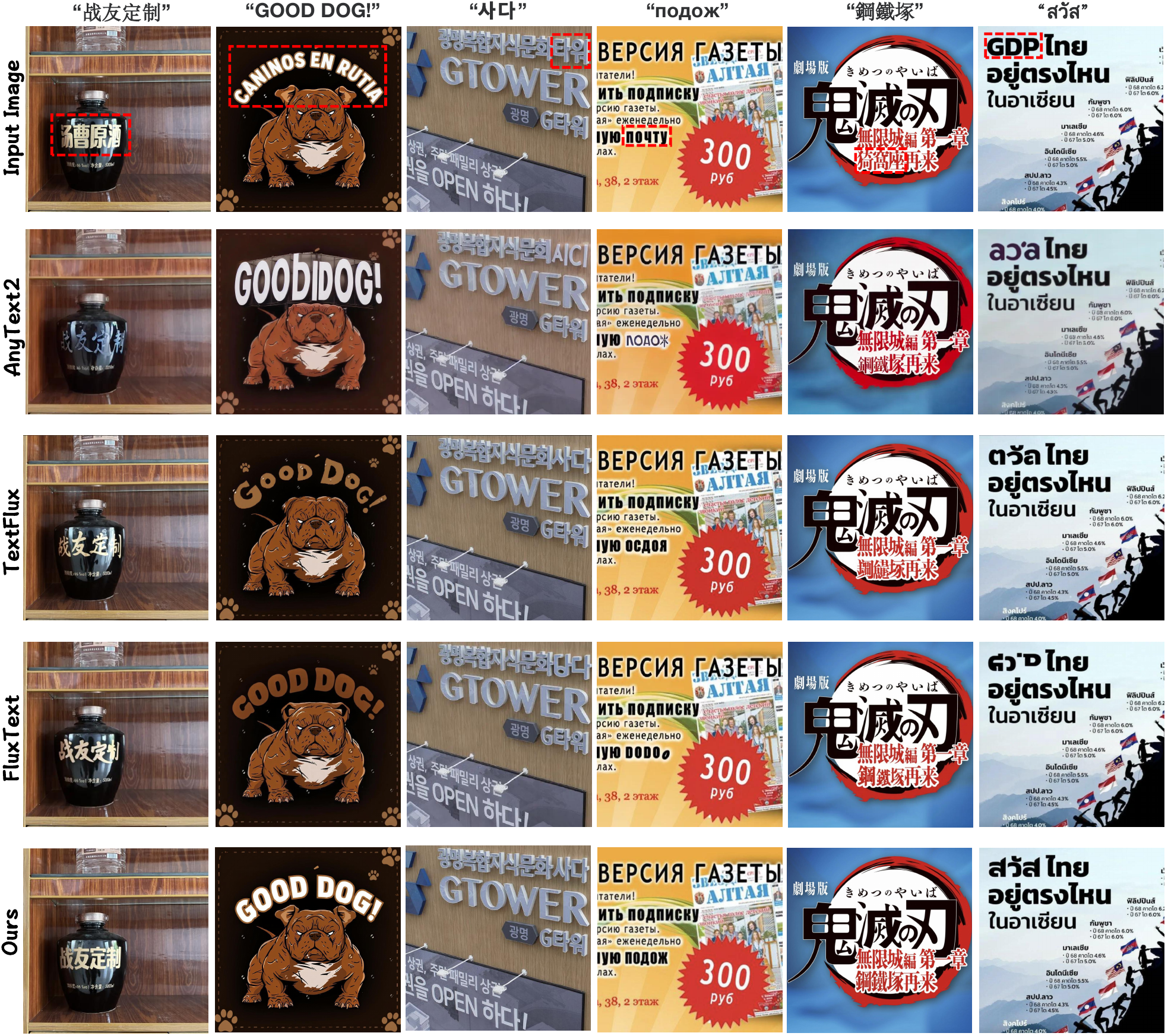}
    \caption{Qualitative results across Chinese, English, Korean, Japanese, Thai, and Russian.}
    \label{fig:quanlitative example}
\end{figure*}
\subsection{Qualitative Result}
\label{subsec:qualitative result}

We conduct a qualitative comparison with representative multilingual scene text editing methods, including AnyText2~\cite{tuo2024anytext2}, TextFlux~\cite{xie2025textflux}, and FluxText~\cite{flux2024}. 

As shown in Figure~\ref{fig:quanlitative example}, our method consistently generates accurate target text across different scripts, preserving correct character structures and visual consistency.
More specifically, results on Russian, Japanese, and Thai demonstrate the effectiveness of our method on long-tail languages with complex and diverse writing systems. In particular, our approach better preserves the original text style in the “\begin{CJK}{UTF8}{gbsn}战友定制\end{CJK}” example, and remains robust in the “GOOD DOG!” case, where the masked region extends beyond the target text, introducing minimal unintended modifications to non-target regions. 
These observations indicate that our method generalizes well across languages while enabling precise and controlled text editing under challenging conditions.

\section{Conclusion}
\label{sec:conclusion}

This paper presents a self-prompting diffusion transformer framework for open-vocabulary scene text editing that preserves both textual correctness and visual style consistency by exploiting in-context information from the original image. 
By directly constructing glyph and style prompts from the input image, the proposed approach enables coherent text generation that remains faithful to the surrounding font appearance, layout, and visual context, without introducing additional glyph or style encoders.
Extensive experiments on large-scale multilingual benchmarks demonstrate that our approach achieves strong performance across diverse languages and scripts, while maintaining robustness under progressive language expansion without negative transfer.
Overall, this paper provides a unified and scalable solution for multilingual scene text editing, and establishes a foundation for extending style-consistent editing to more complex and diverse real-world scenarios.


\section*{Impact Statement}
This work studies scene text editing, which aims to modify textual content in images while preserving visual style and background consistency. The proposed method advances the flexibility and robustness of text editing across languages and scripts, which may benefit applications such as graphic design, content creation, and multilingual visual communication. 

At the same time, like other image editing and generative techniques, scene text editing could be misused to alter visual content in misleading or deceptive ways. This work is intended for research purposes, and we encourage responsible use in accordance with existing ethical guidelines for generative models and visual media editing.

\bibliography{example_paper}
\bibliographystyle{icml2026}

\newpage
\appendix
\onecolumn

\section*{Supplementary Material}
\section{Full Numeric Results}
\label{sup:mstedit}
\subsection{Comparison with General Image Editing Models}

We additionally compare our method with several recent general-purpose image editing models, including Qwen-Image-Edit~\cite{wu2025qwen}, Longcat-Image-Edit~\cite{team2025longcat}, and FireRed-Image-Edit~\cite{team2026firered}. Since these methods are designed for instruction-based editing rather than mask-guided editing, direct comparison is not entirely straightforward. To enable evaluation, we convert our task into an instruction-based setting using prompts such as: ``replace the original text `XXX' with `YYY'.''

\begin{table*}[h]
\centering
\small
\setlength{\tabcolsep}{4pt}
\renewcommand{\arraystretch}{1.08}

\begin{tabular}{lcccc|cccc}
\toprule
\multirow{2}{*}{\textbf{Method}} 
& \multicolumn{4}{c|}{\textbf{English}} 
& \multicolumn{4}{c}{\textbf{Chinese}} \\
\cmidrule(lr){2-5} \cmidrule(lr){6-9}
& Seq.\ Acc & NED & FID & LPIPS 
& Seq.\ Acc & NED & FID & LPIPS \\
\midrule

Qwen-Image-Edit-2509 
& 0.6902 & 0.7422 & 24.20 & 0.1985 
& 0.6288 & 0.6982 & 26.02 & 0.1985 \\

Qwen-Image-Edit-2511 
& 0.7082 & 0.7893 & 23.95 & 0.1874 
& 0.6623 & 0.7102 & 24.29 & 0.1832 \\

Longcat-Image-Edit   
& 0.6725 & 0.7210 & 25.60 & 0.2050 
& 0.6015 & 0.6750 & 27.80 & 0.2070 \\

FireRed-Image-Edit-1.1 
& 0.7375 & 0.8125 & 21.80 & 0.1720 
& 0.6880 & 0.7355 & 22.90 & 0.1705 \\

\textbf{Ours} 
& \textbf{0.8857} & \textbf{0.9568} & \textbf{7.62} & \textbf{0.0365} 
& \textbf{0.8249} & \textbf{0.9147} & \textbf{7.95} & \textbf{0.0268} \\

\bottomrule
\end{tabular}

\caption{Comparison with general image editing models on English and Chinese benchmarks.}
\label{tab:general_edit_comparison}
\end{table*}
Table~\ref{tab:general_edit_comparison} show a consistent ranking across different settings:$\text{Ours} > \text{FireRed} > \text{Qwen-2511} > \text{Qwen-2509} > \text{LongCat}.$

Our method achieves superior performance in text accuracy, localization, and stylistic consistency. We attribute this advantage mainly to the explicit mask-guided formulation, which directly constrains the editable region and avoids the implicit region inference required by instruction-based methods. In contrast, general editing models often suffer from localization and semantic errors, such as editing incorrect regions, incomplete replacement, unintended synonym substitution, or failure to preserve the original typography and layout. These issues become more evident in multilingual scene text editing, where accurate glyph rendering and precise spatial control are critical.

\subsection{Complete Experimental Results on the MSTEdit Dataset}

The complete experimental results corresponding to Figure~\ref{fig:mst-edit result} are presented in Table~\ref{tab:multilingual_results}. 
We additionally include OCR-based baselines, including AnyText2 and TextCtrl. However, these methods are not specifically designed for multilingual scene text editing, and their OCR modules mainly support Chinese and English text, resulting in relatively limited performance on broader multilingual editing scenarios.

\begin{table*}[!h]
\centering
\small
\setlength{\tabcolsep}{4pt}
\renewcommand{\arraystretch}{1.1}
\begin{tabular*}{0.85\textwidth}{@{\extracolsep{\fill}}ccccccccccc}
\toprule
\multirow{2}{*}{\textbf{Language}} 
& \multicolumn{2}{c}{\textbf{AnyText2}} 
& \multicolumn{2}{c}{\textbf{TextCtrl}} 
& \multicolumn{2}{c}{\textbf{FluxText}} 
& \multicolumn{2}{c}{\textbf{TextFlux}} 
& \multicolumn{2}{c}{\textbf{Ours}} \\
\cmidrule(lr){2-3} \cmidrule(lr){4-5} \cmidrule(lr){6-7} \cmidrule(lr){8-9} \cmidrule(lr){10-11}
& Seq. ACC & FID & Seq. ACC & FID & Seq. ACC & FID & Seq. ACC & FID & Seq. ACC & FID \\
\midrule
Arabic   & 0.108 & 75.2 & 0.110 & 74.9 & 0.698 & 33.5 & \underline{0.705} & \underline{30.0} & \textbf{0.762} & \textbf{23.0} \\
Japanese & 0.074 & 78.8 & 0.073 & 79.1 & 0.604 & 32.0 & \underline{0.610} & \underline{29.0} & \textbf{0.670} & \textbf{22.0} \\
Korean   & 0.086 & 80.2 & 0.088 & 79.8 & 0.602 & 35.0 & \underline{0.610} & \underline{31.5} & \textbf{0.678} & \textbf{24.0} \\
French   & 0.227 & 70.1 & 0.226 & 71.4 & \underline{0.735} & \underline{32.5} & 0.744 & 29.8 & \textbf{0.808} & \textbf{22.0} \\
German   & 0.216 & 71.6 & 0.218 & 72.3 & \underline{0.742} & 34.5 & 0.738 & \underline{30.8} & \textbf{0.805} & \textbf{22.5} \\
Italian  & 0.220 & 70.5 & 0.219 & 71.7 & 0.730 & \underline{31.0} & \underline{0.735} & 28.5 & \textbf{0.800} & \textbf{21.5} \\
Bengali  & 0.093 & 80.7 & 0.094 & 80.4 & \underline{0.708} & 34.0 & 0.710 & \underline{30.2} & \textbf{0.785} & \textbf{23.3} \\
Hindi    & 0.100 & 76.2 & 0.099 & 76.4 & 0.702 & 30.5 & \underline{0.713} & \underline{27.8} & \textbf{0.790} & \textbf{21.0} \\
Russian  & 0.087 & 78.4 & 0.089 & 78.1 & \underline{0.705} & 34.0 & 0.708 & \underline{30.9} & \textbf{0.775} & \textbf{24.5} \\
Thai     & 0.061 & 81.5 & 0.060 & 81.8 & 0.588 & 32.0 & \underline{0.595} & \underline{29.0} & \textbf{0.672} & \textbf{22.0} \\
Swahili  & 0.204 & 75.2 & 0.206 & 76.9 & 0.700 & 36.0 & \underline{0.706} & \underline{32.0} & \textbf{0.772} & \textbf{25.0} \\
\midrule
\textbf{AVG} 
& 0.0978 & 77.9 
& 0.0984 & 77.9 
& \underline{0.68} & 33.2 
& 0.69 & \underline{29.6} 
& \textbf{0.76} & \textbf{22.8} \\
\bottomrule
\end{tabular*}
\caption{Multilingual text editing performance comparison across different methods.}
\label{tab:multilingual_results}
\end{table*}

The complete experimental results corresponding to Figure~\ref{fig:ablation language} are presented in Table~\ref{tab:sequential_lang}. Following the predefined language list, we progressively introduce new language data in a cyclic order during training.

\begin{table*}[t]
\centering
\scriptsize
\setlength{\tabcolsep}{2.5pt}
\renewcommand{\arraystretch}{1.0}

\caption{Seq. ACC of initially learned languages under incremental multilingual training. 
Languages are added sequentially in the following order: 
\textit{Arabic $\rightarrow$ English $\rightarrow$ French $\rightarrow$ Chinese $\rightarrow$ German $\rightarrow$ Korean $\rightarrow$ Japanese $\rightarrow$ Italian $\rightarrow$ Bengali $\rightarrow$ Hindi $\rightarrow$ Russian $\rightarrow$ Thai $\rightarrow$ Swahili}. 
Each column denotes the performance after introducing an additional language.
}
\label{tab:sequential_lang}

\resizebox{0.8\textwidth}{!}{
\begin{tabular}{lccccccccccc}
\toprule

\multirow{2}{*}{\textbf{Start}}
& \multicolumn{11}{c}{\textbf{Number of Languages Seen (Sequential Steps)}} \\

\cmidrule(lr){2-12}

& \textbf{1} & \textbf{2} & \textbf{3} & \textbf{4} & \textbf{5} 
& \textbf{6} & \textbf{7} & \textbf{8} & \textbf{9} 
& \textbf{10} & \textbf{11} \\

\midrule

Arabic   
& 0.6812 & 0.6987 & 0.7121 & 0.7094 & 0.7113 
& 0.7215 & 0.7198 & 0.7517 & 0.7589 
& 0.7612 & 0.7596 \\

French   
& 0.7423 & 0.7491 & 0.7486 & 0.7615 & 0.7592 
& 0.7713 & 0.7689 & 0.7917 & 0.7888 
& 0.7909 & 0.8124 \\

German   
& 0.7096 & 0.7112 & 0.7089 & 0.7123 & 0.7216 
& 0.7288 & 0.7321 & 0.7819 & 0.7897 
& 0.8014 & 0.8098 \\

Korean   
& 0.5931 & 0.6017 & 0.6024 & 0.6392 & 0.6415 
& 0.6387 & 0.6511 & 0.6623 & 0.6589 
& 0.6698 & 0.6812 \\

Japanese 
& 0.6094 & 0.6321 & 0.6313 & 0.6422 & 0.6389 
& 0.6415 & 0.6398 & 0.6427 & 0.6411 
& 0.6589 & 0.6713 \\

Italian  
& 0.7315 & 0.7398 & 0.7422 & 0.7489 & 0.7716 
& 0.7687 & 0.7812 & 0.7794 & 0.7821 
& 0.7986 & 0.8015 \\

Bengali  
& 0.7089 & 0.7115 & 0.7312 & 0.7298 & 0.7421 
& 0.7389 & 0.7417 & 0.7719 & 0.7794 
& 0.7812 & 0.7786 \\

Hindi    
& 0.6917 & 0.6994 & 0.7215 & 0.7189 & 0.7486 
& 0.7521 & 0.7589 & 0.7697 & 0.7723 
& 0.7688 & 0.7912 \\

Russian  
& 0.7113 & 0.7098 & 0.7124 & 0.7391 & 0.7415 
& 0.7389 & 0.7422 & 0.7398 & 0.7411 
& 0.7617 & 0.7689 \\

Thai     
& 0.5715 & 0.5898 & 0.5912 & 0.6124 & 0.6189 
& 0.6215 & 0.6197 & 0.6589 & 0.6612 
& 0.6578 & 0.6694 \\

Swahili  
& 0.6792 & 0.6815 & 0.6789 & 0.7012 & 0.6994 
& 0.7117 & 0.7398 & 0.7689 & 0.7713 
& 0.7694 & 0.7812 \\

\bottomrule
\end{tabular}
}
\end{table*}
\subsection{Detailed ablation of glyph prompt \& style prompt \& cooldown}

We conduct an ablation study to evaluate different prompt setting, including \textit{text-only}, \textit{text+glyph}, \textit{text+style}, and \textit{text+glyph+style}, across two training stages: pretraining and cooldown. Experiments are performed on both English and Chinese benchmarks. In the cooldown stage, we further compare the effects of paired and unpaired supervision.

The results on Table~\ref{tab:ablation} can be summarized as follows: (1) The text-only setting performs the worst, as no structural or visual reference is provided. (2) In pretraining, text+glyph achieves the best performance, while adding the style prompt degrades results, likely because the source and target texts are identical, which leads to overfitting and copying behavior. (3) In the cooldown stage, text+glyph under both w/o paired and w/ paired yields comparable performance, indicating a continuation of pretraining; text+style provides limited gains due to missing structural guidance, whereas combining glyph and style with paired data yields the most significant improvement. 

These findings motivate our strategy of glyph-only prompt on self-supervised data, followed by joint glyph and style prompt with paired data in the cooldown stage.
\begin{table}[!h]
\centering
\scriptsize
\setlength{\tabcolsep}{2.5pt}
\renewcommand{\arraystretch}{1.05}
\resizebox{0.85\linewidth}{!}{
\begin{tabular}{c|cc|cc|cc|cc}
\toprule
\multirow{2}{*}{\textbf{Language}} 
& \multicolumn{2}{c|}{\multirow{2}{*}{\textbf{Setting}}}
& \multicolumn{2}{c|}{\multirow{2}{*}{\textbf{Pretrain Stage}}}
& \multicolumn{4}{c}{\textbf{Cooldown Stage}} \\
\cmidrule(lr){6-9}
& \multicolumn{2}{c|}{} 
& \multicolumn{2}{c|}{} 
& \multicolumn{2}{c|}{w/o paired}
& \multicolumn{2}{c}{w/ paired} \\
\cmidrule(lr){2-3} 
\cmidrule(lr){4-5} 
\cmidrule(lr){6-7} 
\cmidrule(lr){8-9}

& Glyph & Style 
& Seq. ACC & FID 
& Seq. ACC & FID 
& Seq. ACC & FID \\
\midrule

\multirow{4}{*}{\textbf{English}}

& $\times$ & $\times$ 
& 0.4127 & 28.73 
& - & - 
& - & - \\

& $\checkmark$ & $\times$ 
& 0.8738 & 11.56 
& 0.8788{\scriptsize\color{green}(+0.005)} 
& 11.52{\scriptsize\color{green}(+0.040)} 
& 0.8785{\scriptsize\color{green}(+0.005)} 
& 11.55{\scriptsize\color{green}(+0.010)} \\

& $\times$ & $\checkmark$ 
& 0.5236 & 22.48 
& 0.5243{\scriptsize\color{green}(+0.001)} 
& 22.48{\scriptsize\color{red}(-0.001)} 
& 0.5248{\scriptsize\color{green}(+0.001)} 
& 22.48{\scriptsize\color{red}(-0.001)} \\

& $\checkmark$ & $\checkmark$ 
& 0.5412 & 18.20 
& 0.5440{\scriptsize\color{green}(+0.003)} 
& 18.22{\scriptsize\color{red}(-0.019)} 
& 0.7982{\scriptsize\color{green}(+0.257)} 
& 10.69{\scriptsize\color{green}(+7.511)} \\

\midrule

\multirow{4}{*}{\textbf{Chinese}}

& $\times$ & $\times$ 
& 0.3563 & 31.28 
& - & - 
& - & - \\

& $\checkmark$ & $\times$ 
& 0.8125 & 12.01 
& 0.8176{\scriptsize\color{green}(+0.005)} 
& 11.97{\scriptsize\color{green}(+0.040)} 
& 0.8169{\scriptsize\color{green}(+0.004)} 
& 12.01{\scriptsize\color{green}(+0.000)} \\

& $\times$ & $\checkmark$ 
& 0.4687 & 24.16 
& 0.4678{\scriptsize\color{red}(-0.001)} 
& 24.19{\scriptsize\color{red}(-0.030)} 
& 0.4695{\scriptsize\color{green}(+0.001)} 
& 24.12{\scriptsize\color{green}(+0.040)} \\

& $\checkmark$ & $\checkmark$ 
& 0.4879 & 19.43 
& 0.4911{\scriptsize\color{green}(+0.003)} 
& 19.47{\scriptsize\color{red}(-0.040)} 
& 0.7426{\scriptsize\color{green}(+0.255)} 
& 11.82{\scriptsize\color{green}(+7.610)} \\

\bottomrule
\end{tabular}
}

\caption{
Ablation study of glyph and style prompts on English and Chinese benchmarks.
Glyph and Style denote the glyph and style prompts, respectively. 
w/o paired indicates training with self-supervised (unpaired) data, while 
w/ paired denotes training with paired data.
}
\label{tab:ablation}
\end{table}

\section{Visualization Results}

\subsection{Out-of-vocabulary (OOV) Instances}

\begin{figure}[h]
    \centering
    \includegraphics[width=\textwidth]{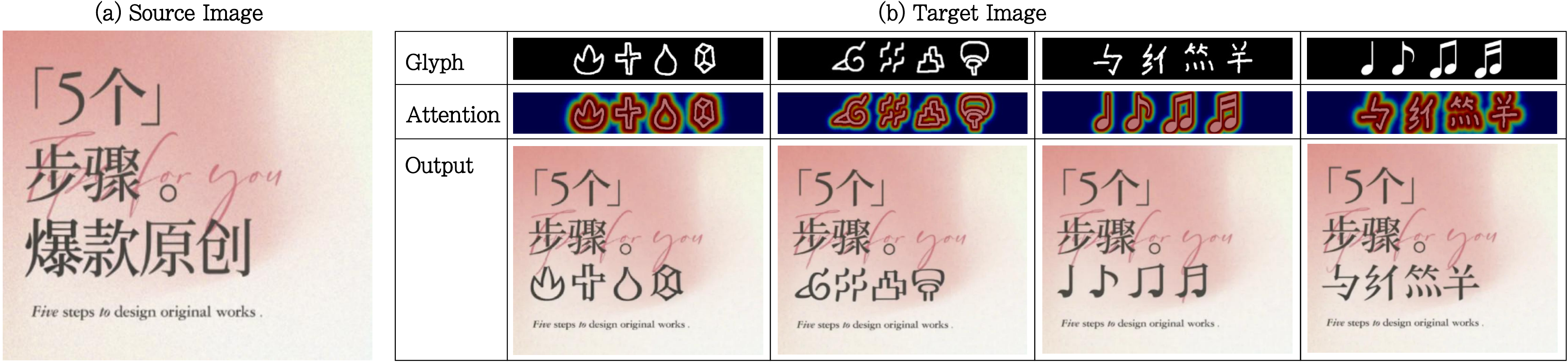}
    \caption{Results of manually designed stroke editing.}
    \label{fig:stroke result}
\end{figure}

\textbf{Manually design characters.} 
We further manually design a set of synthetic symbol characters that do not exist in real-world writing systems and use them as target texts for evaluation. 
Figure~\ref{fig:stroke result} show that our method still achieves visually coherent and structurally plausible editing results on these previously unseen characters. 
This observation suggests that our method primarily learns stroke-level compositional representations rather than relying solely on character-level memorization. Consequently, even for non-existent characters, the model can still generate reasonable glyph structures as long as they share common stroke patterns with existing characters.

\begin{figure}[!h]
    \centering
    \includegraphics[width=\textwidth]{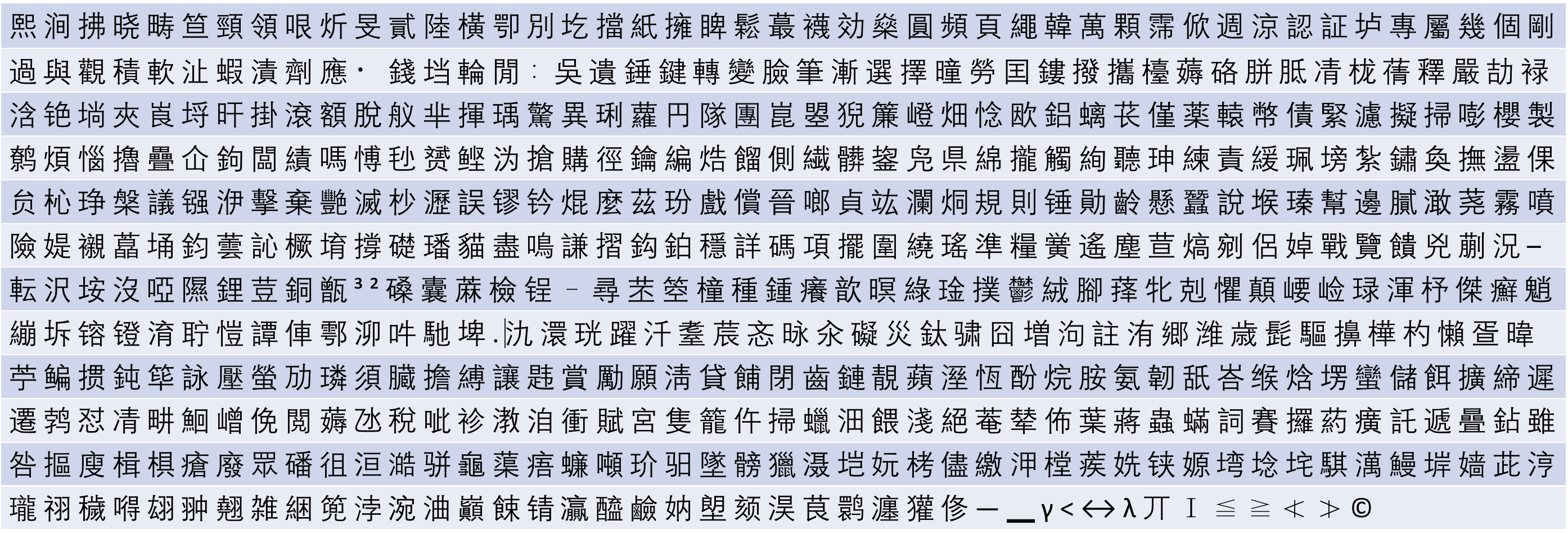}
    \caption{Out-of-vocabulary character set.}
    \label{fig:oov set}
\end{figure}

\begin{figure}[!h]
    \centering
    \includegraphics[height=0.65\textheight]{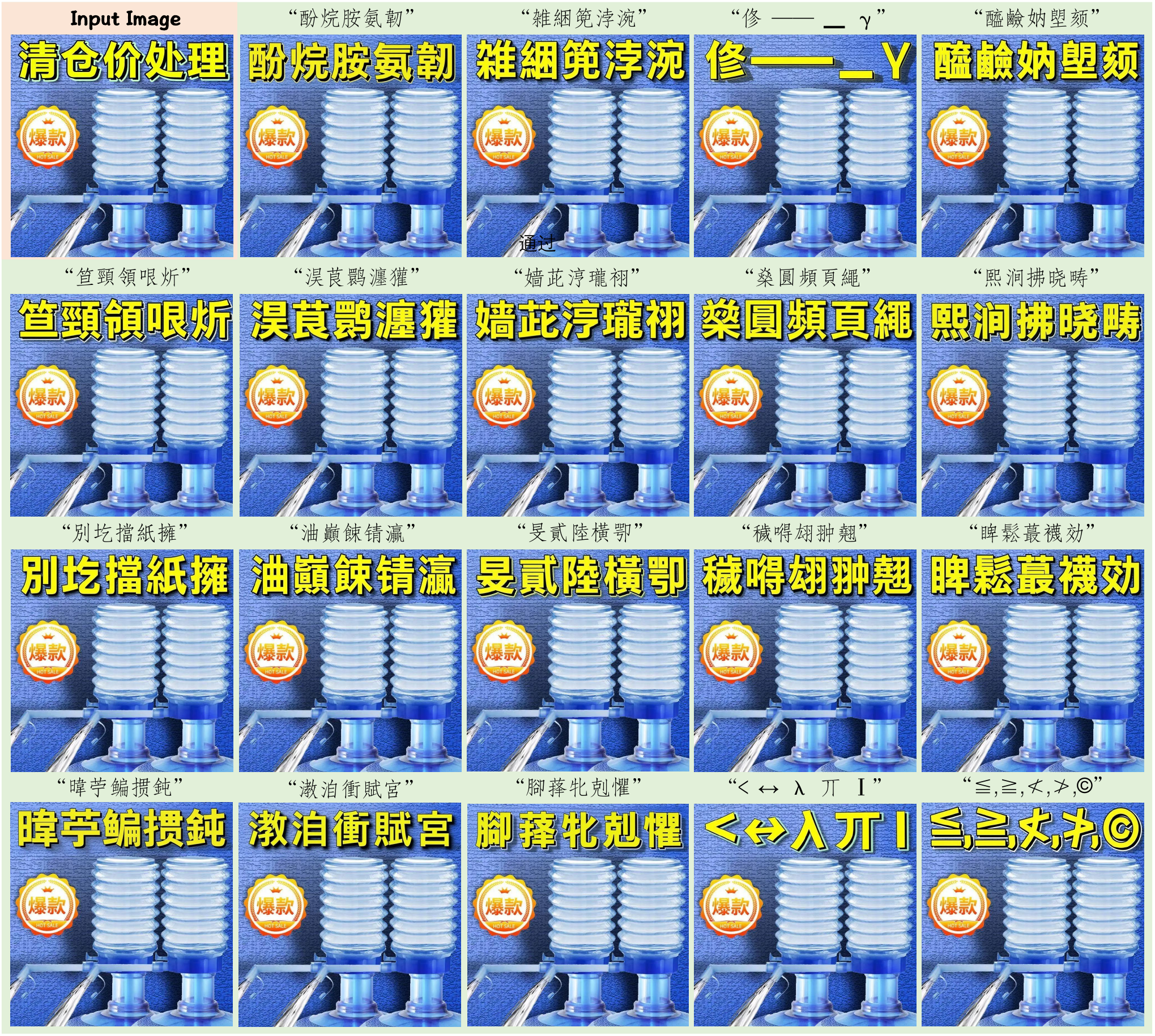}
    \caption{Out-of-vocabulary text editing results. The top-left panel shows the input image, where the original text reads ''\begin{CJK*}{UTF8}{gbsn}清仓价处理\end{CJK*}''.}
    \label{fig:oov result}
\end{figure}

\textbf{Out-of-vocabulary (OOV) characters.}
We collect 537 rare characters that are not supported by standard OCR vocabularies and conduct a dedicated evaluation on these OOV cases. The complete OOV vocabulary collected in this work is illustrated in Figure~\ref{fig:oov set}. Figure~\ref{fig:oov result} results demonstrate that our method still maintains strong performance on the OOV vocabulary, even though these characters rarely appear in existing datasets.

\clearpage

\subsection{Multilingual Text Editing}

\begin{figure}[!h]
    \centering
    \includegraphics[height=0.97\textheight,keepaspectratio]{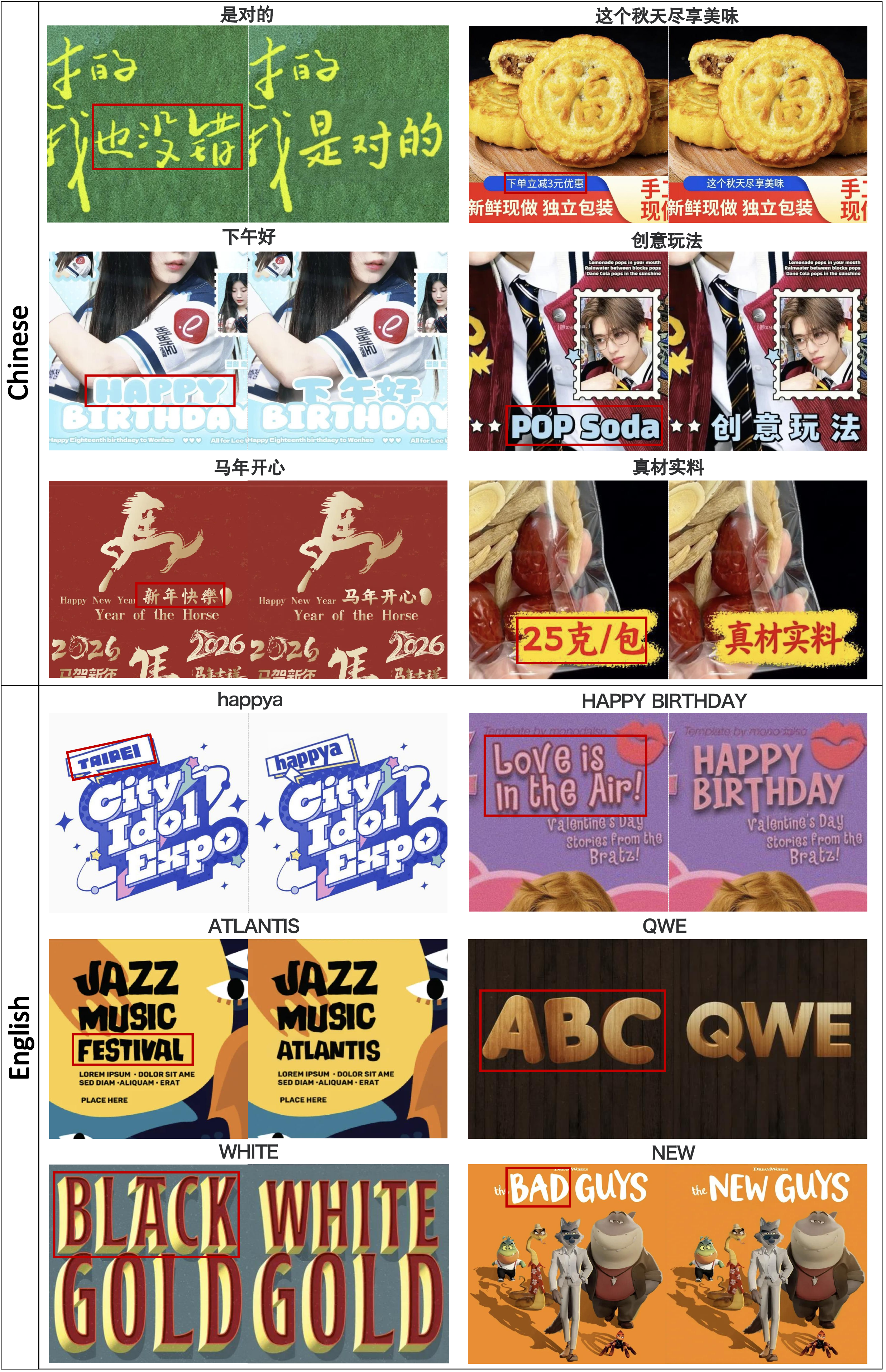}
    \caption{Qualitative results on chinese and english text editing.}
    \label{fig:zh-en}
\end{figure}

\begin{figure}[!h]
    \centering
    \includegraphics[height=0.97\textheight,keepaspectratio]{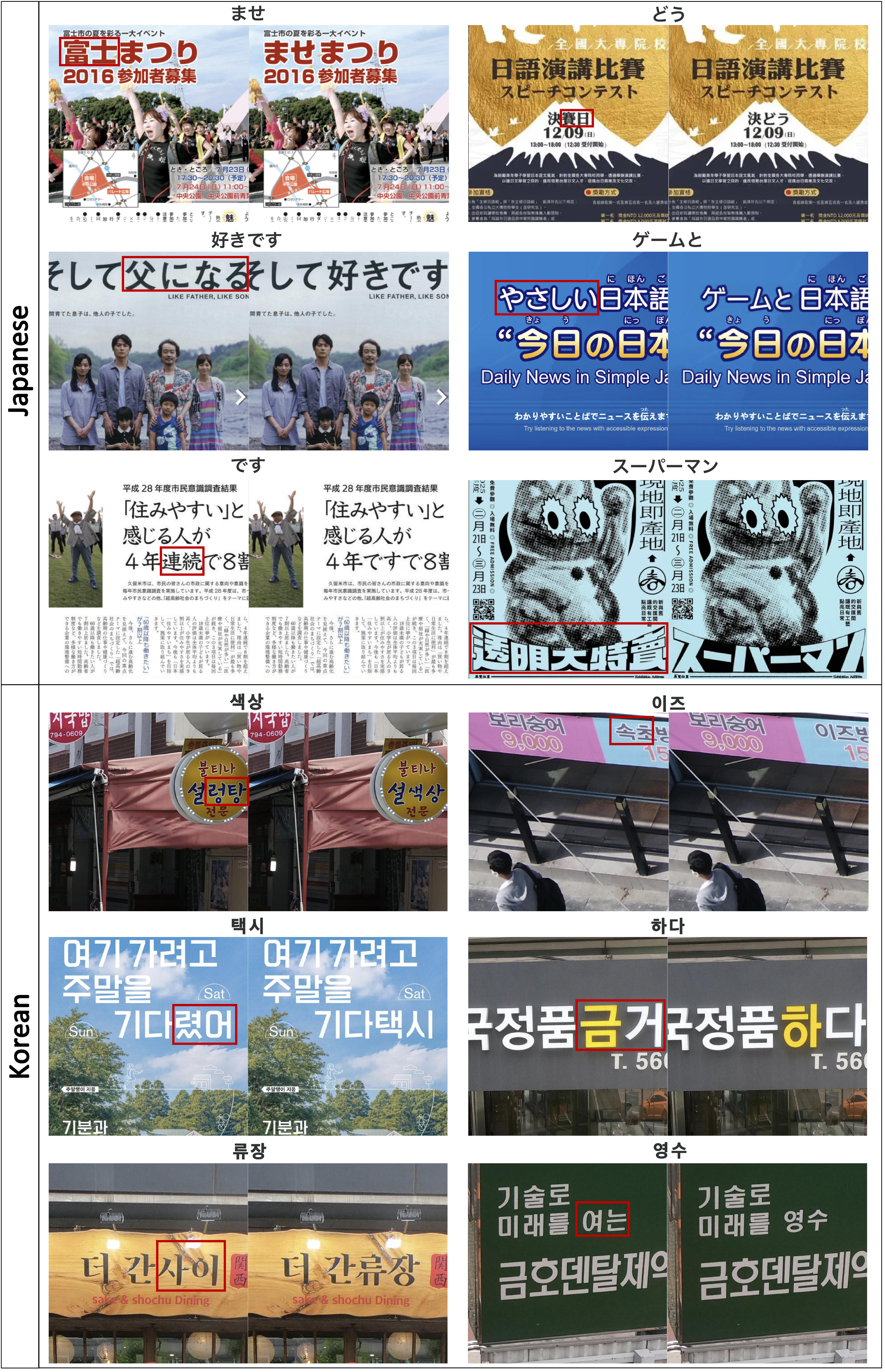}
    \caption{Qualitative results on japanese and korean text editing.}
    \label{fig:ja-ko}
\end{figure}

\begin{figure}[!h]
    \centering
    \includegraphics[height=0.97\textheight,keepaspectratio]{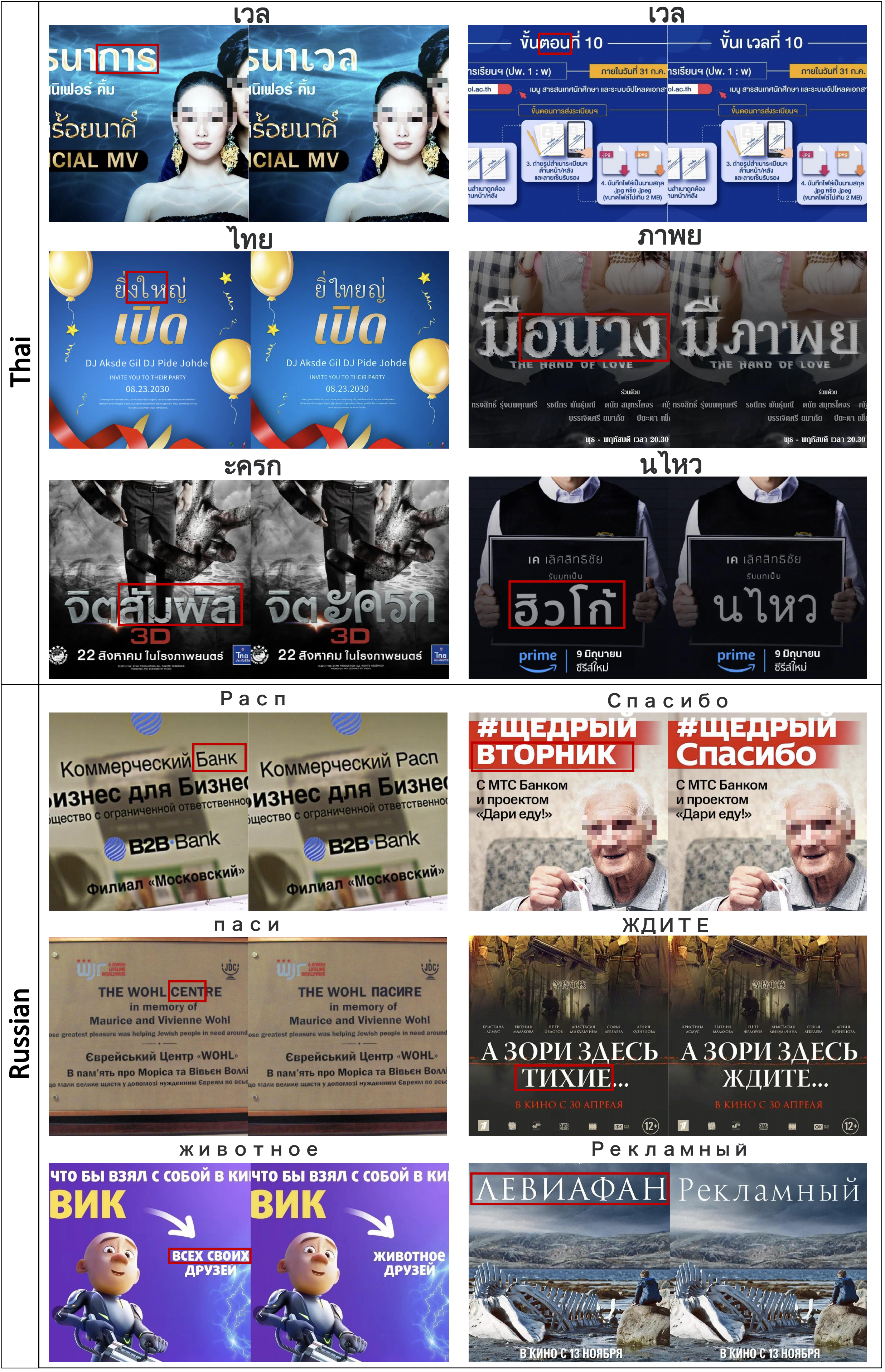}
    \caption{Qualitative results on thai and russian text editing.}
    \label{fig:th-ru}
\end{figure}

For the languages covered in MSTEdit, we provide additional visualization results of our method on several representative language groups, including English/Chinese (Figure~\ref{fig:zh-en}), Japanese/Korean (Figure~\ref{fig:ja-ko}), and Thai/Russian (Figure~\ref{fig:th-ru}). The red boxes indicate the target mask regions for text editing.

\subsection{Performance on long text}

\begin{figure}[!h]
    \centering
    \includegraphics[width=\textwidth]{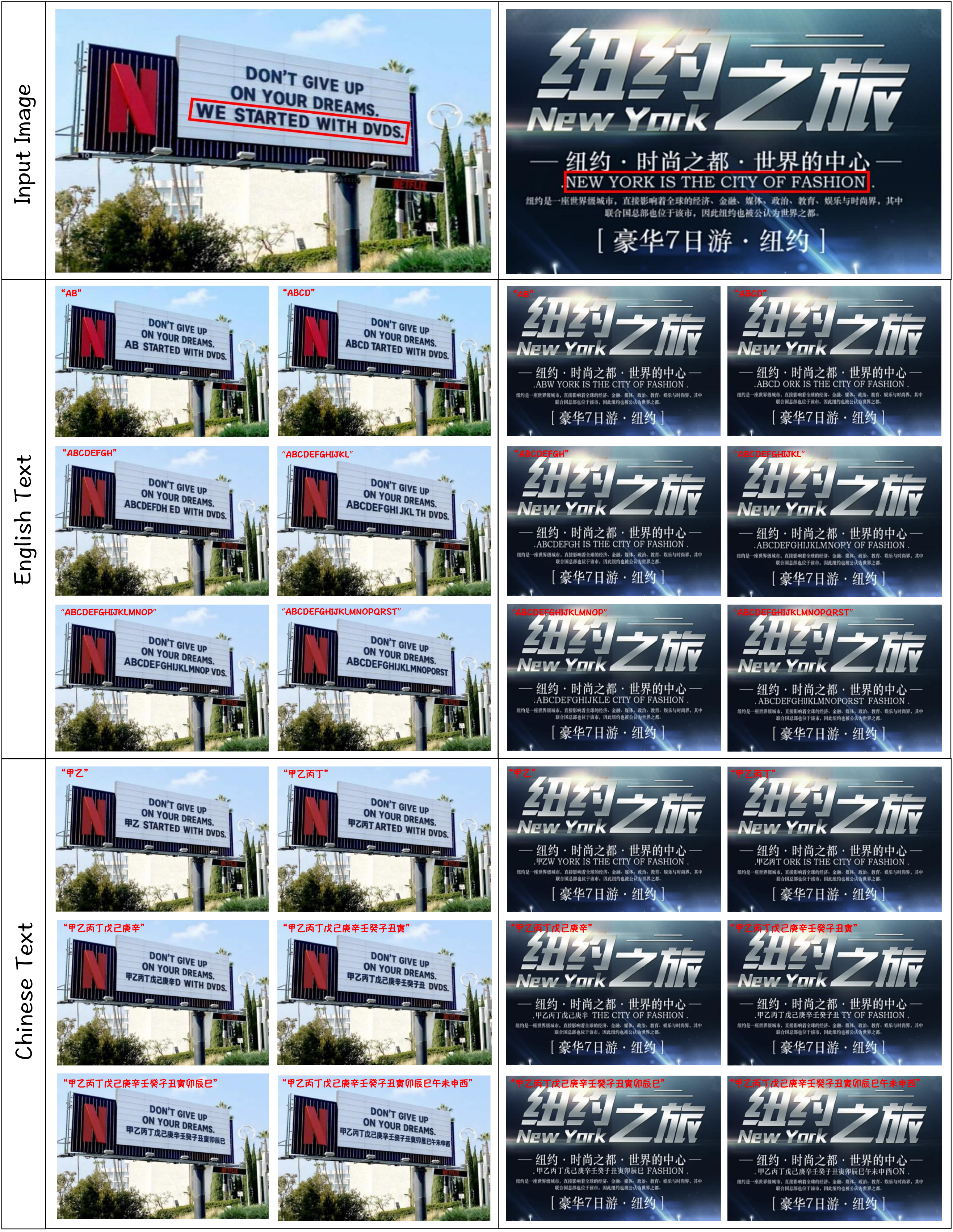}
    \caption{Text editing results under different text lengths.}
    \label{fig:long text}
\end{figure}

Using Chinese and English as representative examples, we present text editing results with text lengths ranging from 2 to 20 characters. As shown in Figure~\ref{fig:long text}, the performance gradually degrades when the text length exceeds 16 characters, mainly reflected in artifacts such as stroke distortion and reduced structural consistency. For even longer text editing scenarios, segmented editing provides a practical solution. Nevertheless, compared with existing approaches, our method still supports substantially longer text editing while maintaining favorable visual quality and textual consistency.

\clearpage
\section{Paired Dataset Construction for Cooldown Training}
\label{sup:dataset}

The paired dataset used in the cooldown stage is constructed using Nano Banana Pro based on real-world images containing at least one visible text instance. For each image, a single target text region is selected and edited to form an original–edited image pair. In total, more than 15,000 candidate pairs are generated, restricted to Chinese and English text editing scenarios.
To ensure high-quality supervision, all generated samples undergo a rigorous human filtering process based on three criteria:

\begin{itemize}[itemsep=0pt, topsep=0pt]
    \item \textbf{Editing correctness:} the source text must be accurately replaced by the target text.
    \item \textbf{Style consistency:} the edited text should preserve the original font, texture, color, and other visual attributes.
    \item \textbf{Content preservation:} all non-target regions should remain unchanged.
\end{itemize}

After filtering, annotators further annotate the edited text regions with bounding boxes, which are used to derive training masks. The final dataset contains 4,000 high-quality image pairs selected from over 15,000 candidates, with more than 11,000 samples discarded during filtering. The entire process involves 10 annotators over approximately two months.

\section{Failure Cases}
The failure cases of our method mainly fall into three scenarios: few-to-many editing, many-to-few editing, and multi-line text editing.

\begin{figure}[!h]
    \centering
    \includegraphics[width=\textwidth]{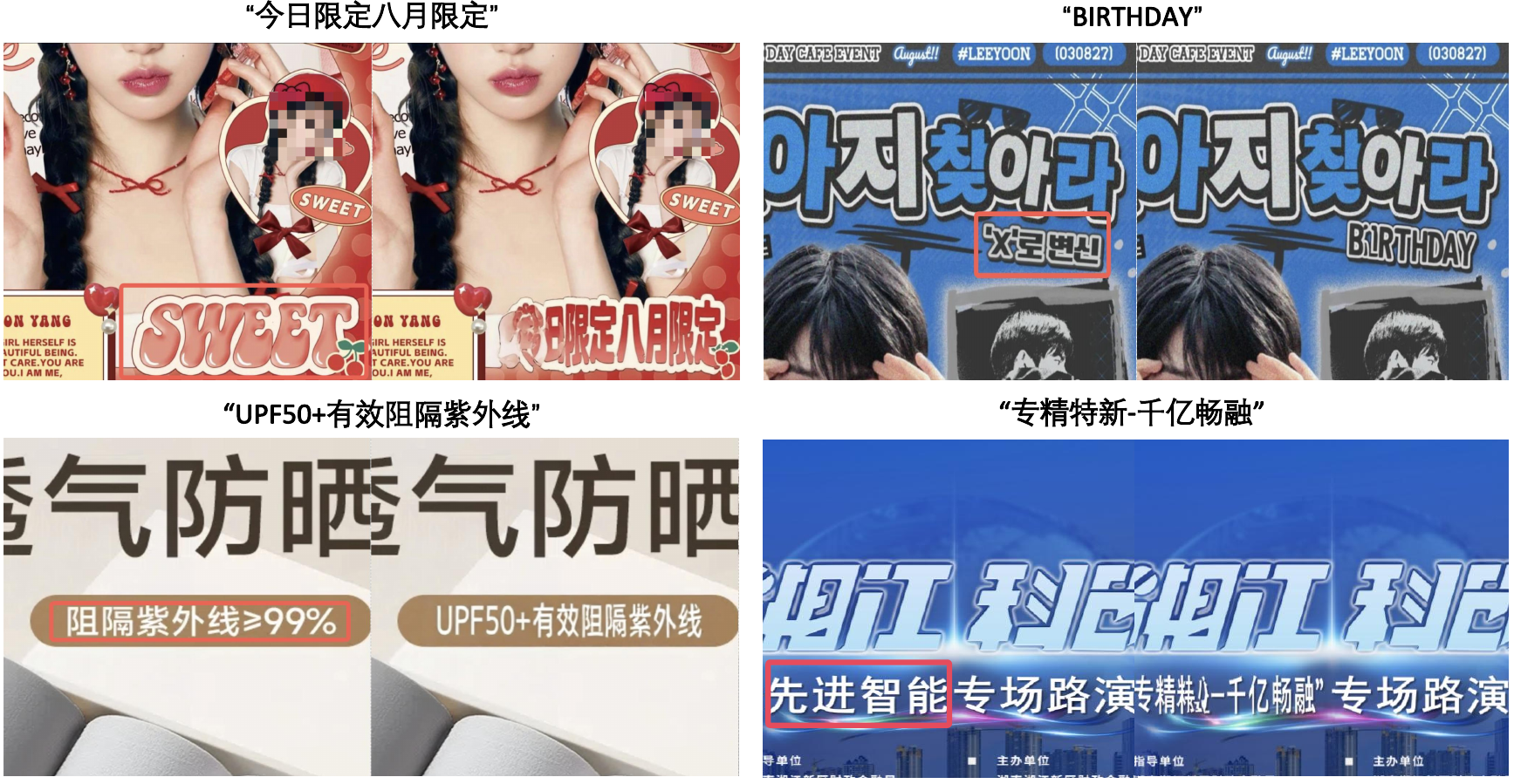}
    \caption{Few-to-many character editing results.}
    \label{fig:few-to-many}
\end{figure}

\begin{figure}[!h]
    \centering
    \includegraphics[width=\textwidth]{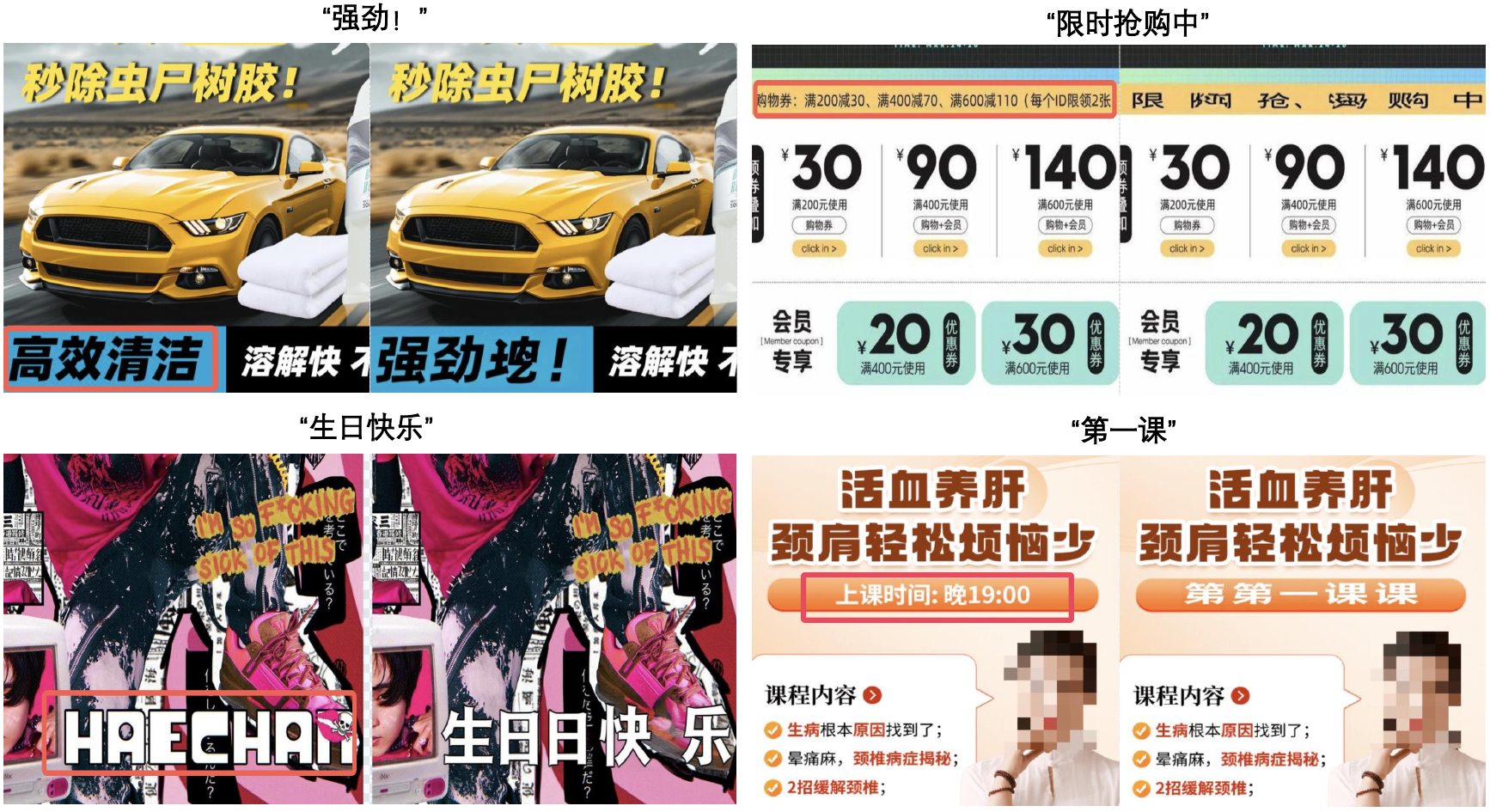}
    \caption{Many-to-few character editing results.}
    \label{fig:many-to-few}
\end{figure}

\begin{figure}[!h]
    \centering
    \includegraphics[width=\textwidth]{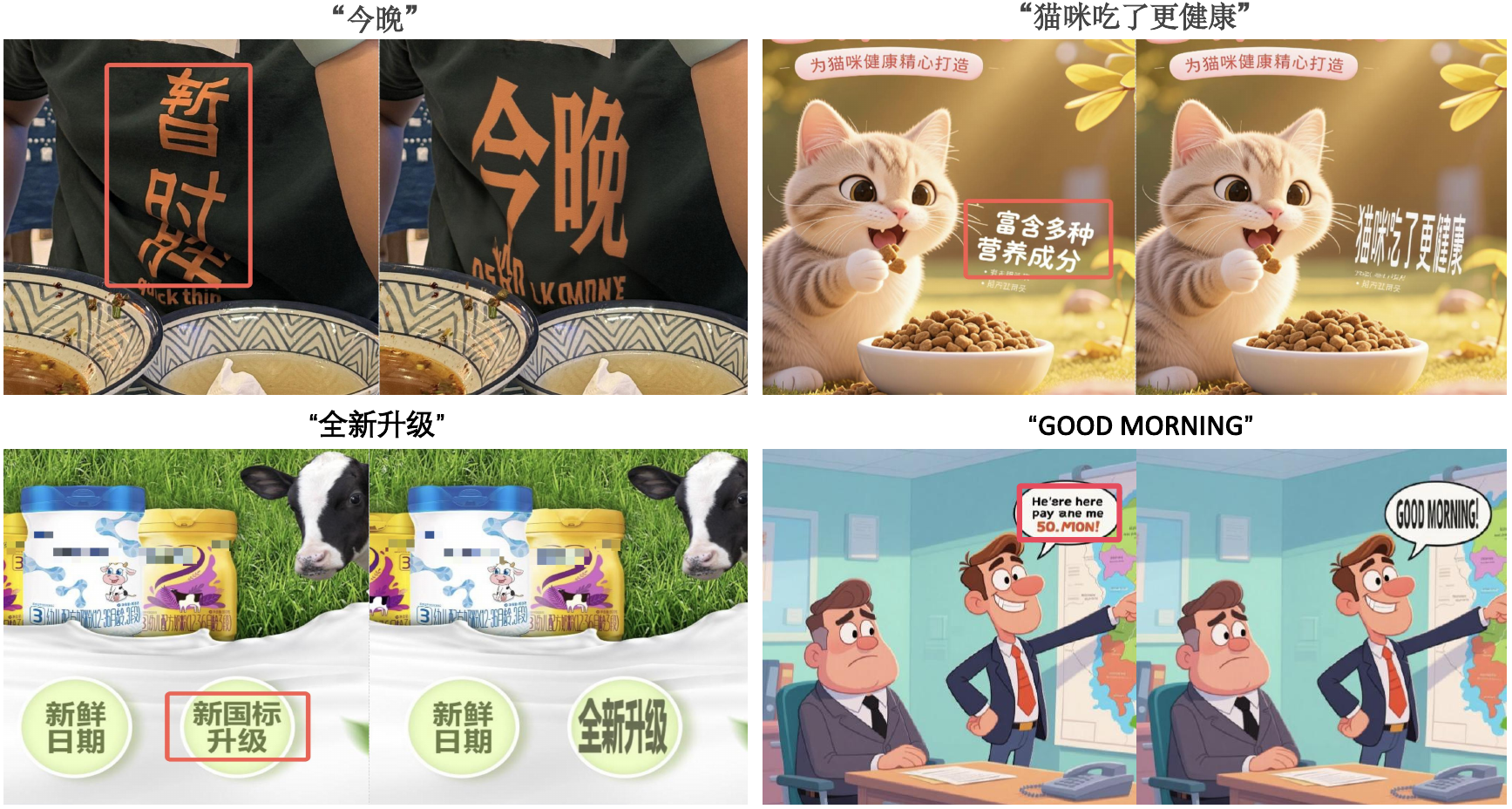}
    \caption{Multi-line text editing results.}
    \label{fig:multi-line}
\end{figure}

\begin{itemize}[itemsep=0pt, topsep=0pt]
    \item \textbf{Few-to-many editing.} When the mask region originally covers only a small amount of text but the target text contains substantially more characters, the generated characters are forced to fit within the limited mask area, leading to artifacts such as character compression and stroke distortion, as illustrated in Figure~\ref{fig:few-to-many}.
    \item \textbf{Many-to-few editing.} When the mask region originally covers a large amount of text but the target text contains significantly fewer characters, excessive blank regions may cause the model to generate hallucinated characters or redundant strokes to adapt to the mask layout, as shown in Figure~\ref{fig:many-to-few}.
    \item \textbf{Multi-line text editing.} When the mask region spans multiple lines of text, the generated results often fail to preserve a proper multi-line layout, as shown in Figure~\ref{fig:multi-line}. This limitation mainly arises because the glyph prompts are rendered in a single-line format, preventing the model from adaptively performing line wrapping during generation.
\end{itemize}


\end{document}